\documentclass[a4paper,fleqn]{cas-sc}
 \usepackage[authoryear]{natbib}

\usepackage{hyperref}
\hypersetup{
	colorlinks=false,
}

\usepackage{url}
\usepackage{graphicx}
\usepackage{wrapfig}
\usepackage{float}
\usepackage{subfig}
\usepackage{stfloats}
\usepackage{epsfig}
\usepackage{makecell}
\usepackage{booktabs} 
\usepackage{multirow}
\usepackage{threeparttable}
\usepackage{rotating} 
\usepackage{array}
\usepackage{algorithm}
\usepackage{algorithmic}
\usepackage{amsmath,amssymb,amsfonts}
\RequirePackage{lineno}

\def\tsc#1{\csdef{#1}{\textsc{\lowercase{#1}}\xspace}}
\tsc{WGM}
\tsc{QE}
\tsc{EP}
\tsc{PMS}
\tsc{BEC}
\tsc{DE}

\begin{document}
	\let\WriteBookmarks\relax
	\def\floatpagepagefraction{1}
	\def\textpagefraction{.001}
	
	\shorttitle{GBT: Two-stage Transformer Framework for Non-stationary Time Series Forecasting}
	
	\shortauthors{Li Shen~et al.}
	
	\title [mode = title]{GBT: Two-stage Transformer Framework for Non-stationary Time Series Forecasting} 
	
	

	%
	\author[1]{Li Shen}
	\author[1]{Yuning Wei}[orcid=0000-0002-5438-4711]
	\cormark[1]	
	\ead{yuning@buaa.edu.cn}

	\author[1]{Yangzhu Wang}
	
	\affiliation[1]{organization={Beihang University}\\Postal Address: RM.807, 8th Dormitory, Dayuncun Residential Quarter, No.29, Zhichun Road
		Beijing 100191
		P.R. China}	
	\cortext[cor1]{Corresponding author}	
\begin{abstract}
	This paper shows that time series forecasting Transformer (TSFT) suffers from severe over-fitting problem caused by improper initialization method of unknown decoder inputs, {especially} when handling non-stationary time series. Based on this observation, we propose \textbf{\emph{GBT}}, a novel two-stage \textbf{\emph{T}}ransformer framework with \textbf{\emph{G}}ood \textbf{\emph{B}}eginning. It decouples the prediction process of TSFT into two stages, including \textit{Auto-Regression} stage and \textit{Self-Regression} stage to tackle the problem of different statistical properties between input and prediction sequences. Prediction results of \textit{Auto-Regression} stage serve as a `\textit{Good Beginning}', i.e., a better initialization for inputs of \textit{Self-Regression} stage. We also propose {the} \textbf{\textit{E}}rror \textbf{\textit{S}}core \textbf{\textit{M}}odification module to further enhance the forecasting capability of the \textit{Self-Regression} stage in GBT. Extensive experiments on seven benchmark datasets demonstrate that GBT outperforms SOTA TSFTs (FEDformer, Pyraformer, ETSformer, etc.) and many other forecasting models (SCINet, N-HiTS, etc.) with only canonical attention and convolution while owning less time and space complexity. It is also general enough to couple with these models to strengthen their forecasting capability. The source code is available at: \url{https://github.com/OrigamiSL/GBT}
\end{abstract}
	
		
\begin{keywords}
	Time Series Forecasting \sep
	Non-Stationary Time Series \sep
	Neural Network \sep
	Transformer
\end{keywords}
		
\maketitle
	
\section{Introduction}
\label{introduction}	
Popular machine learning technology and neural networks {\citep{SSM,TCN,LSTNet, WEERAKODY2021161,HE2022143}} have been applied in time series forecasting of various fields {\citep{zhao2021empirical,ahmad2019human,traffic_flow_forecasting}} more and more widely to tackle complicated forecasting situations. \textbf{\textit{T}}ime \textbf{\textit{S}}eries \textbf{\textit{F}}orecasting \textbf{\textit{T}}ransformer (\textbf{\textit{TSFT}}), which benefits from long-term dependency capturing capability of attention mechanism {\citep{attention2017}}, dominates long-horizon forecasting researches based on neural networks {\citep{informer,logtrans,ETSformer,Pyraformer}} in the past few years. However, some recent researches {\citep{RevIN,non-stationaryTransformer}} point out that TSFTs are weak in tackling non-stationary time series, which are universal in real-world, and will suffer severe over-fitting problem caused by different statistical dynamics of different local windows. Furthermore, some advanced deep forecasting models based on simple MLP (or Linear) {\citep{DLinear, N-HiTS,TimesNet,LightTS}} have reported state-of-the-art forecasting performances which can compete and outperform TSFTs. Therefore, it indicates that there must exist certain deficiencies in TSFT architectures. In this paper, we find out that the \textit{initialization method of decoder inputs} is responsible for it. \par
\begin{figure}[t]
	\centering	
	\subfloat[]{\includegraphics[width=0.3\columnwidth]{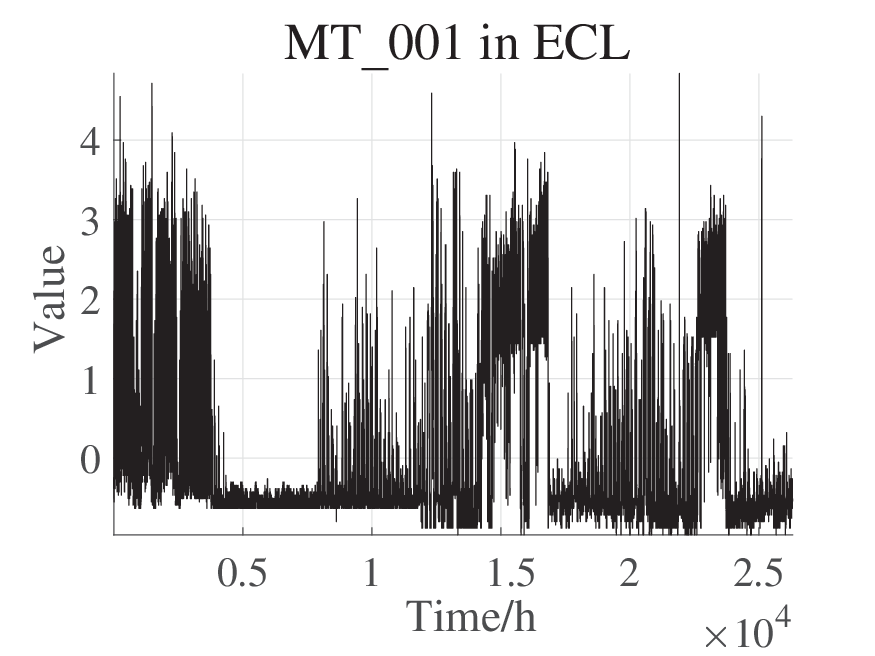}\label{fig1a}}
	\subfloat[]{\includegraphics[width=0.3\columnwidth]{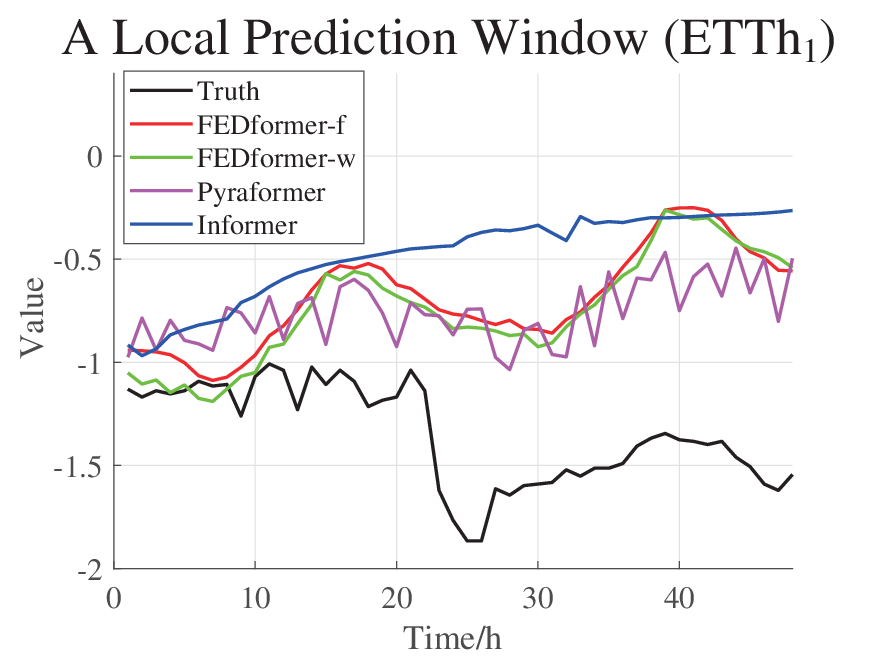}\label{fig1b}}	
	\caption{(a) is the data distribution of `MT\_001' in ECL dataset after Z-score standardization. (b) are prediction results of some TSFTs when handling a local window in ETTh$_1$ dataset. It is obvious that they all suffer severe over-fitting problem.}
	\label{fig1}
\end{figure}

As a significiant component of Transformer architecture, decoder plays the role of inferring target outputs {\citep{BERT,focal}}, which are normally unknown, via representations provided by encoder. In time series forecasting tasks, target outputs will be prediction sequences. As inputs of decoder in TSFTs locate at the unknown timespan, many TSFTs initialize them with zero vector/tensor {\citep{informer,logtrans}}. This initialization method is established upon the assumption that average values of unknown prediction sequences are zero in statistical manner after Z-score standardization {\citep{Autoformer,ETSformer,FEDformer}}, which commonly serves as preprocessing procedure of time series. Z-score standardization forces global data distribution of time series into a specific distribution with average value of zero and variance of one. Therefore, if {the} statistics of time series are invariant with time, i.e., time series is stationary, average values of prediction sequences are supposed to be zero, which means that zero-initialization method is appropriate under stationary conditions. However, this assumption is no longer rational under non-stationary time series. Take a widely-used real-world electricity consumption dataset {\citep{informer,ETSformer,FEDformer}}, whose time series are non-stationary, as an example. It can be observed in Figure \ref{fig1a} that {the} statistics of its sub-sequences keep changing with different timespans though after Z-score standardization so that rarely does sub-sequence of non-stationary time series own {the} statistics of 0-mean and 1-variance. Therefore, if we initialize decoder inputs with zero vector/tensor, TSFTs will certainly suffer from {the} over-fitting problem (Figure \ref{fig1b}), illustrating that zero-initialization method for decoder inputs is not a proper solution in practice.\par
To address the issue of zero-initialization of decoder inputs, some TSFTs additionally employ \textbf{start token} {\citep{informer,logtrans}} or \textbf{decomposed trend terms} of input sequences from encoder {\citep{Autoformer, FEDformer, ETSformer}} to initialize inputs of decoders. As we have analyzed before, statistical properties of local sub-sequences within non-stationary time series are dynamic so that data distributions of input/output sequence are rarely identical in real-world practice. It means that these two methods are not always effective. Meanwhile, they will bring extra computation cost, which is expensive especially considering the quadratic complexity of attention mechanism in Transformer. We will provide more detailed analysis in later Section \ref{Methodology}.\par
\begin{figure}[t]
	\centering
	\subfloat[]{\includegraphics[width=0.34\columnwidth]{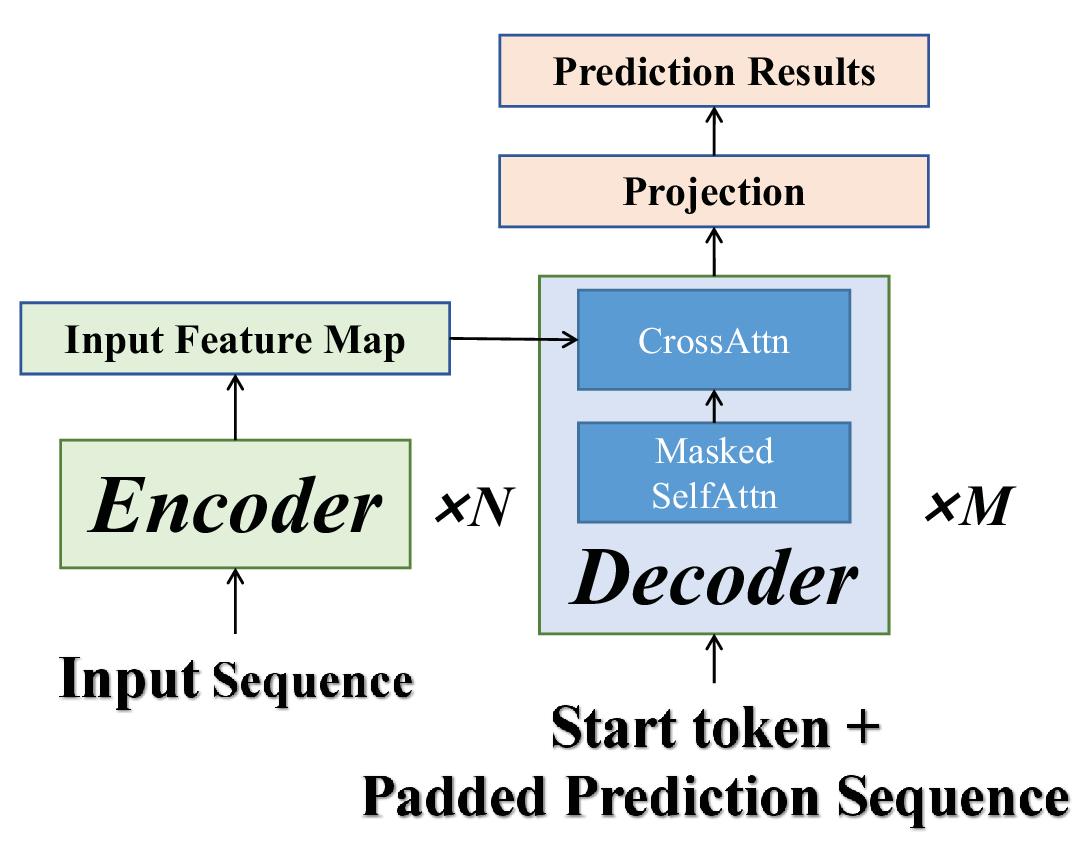}\label{fig2a}}
	\subfloat[]{\includegraphics[width=0.34\columnwidth]{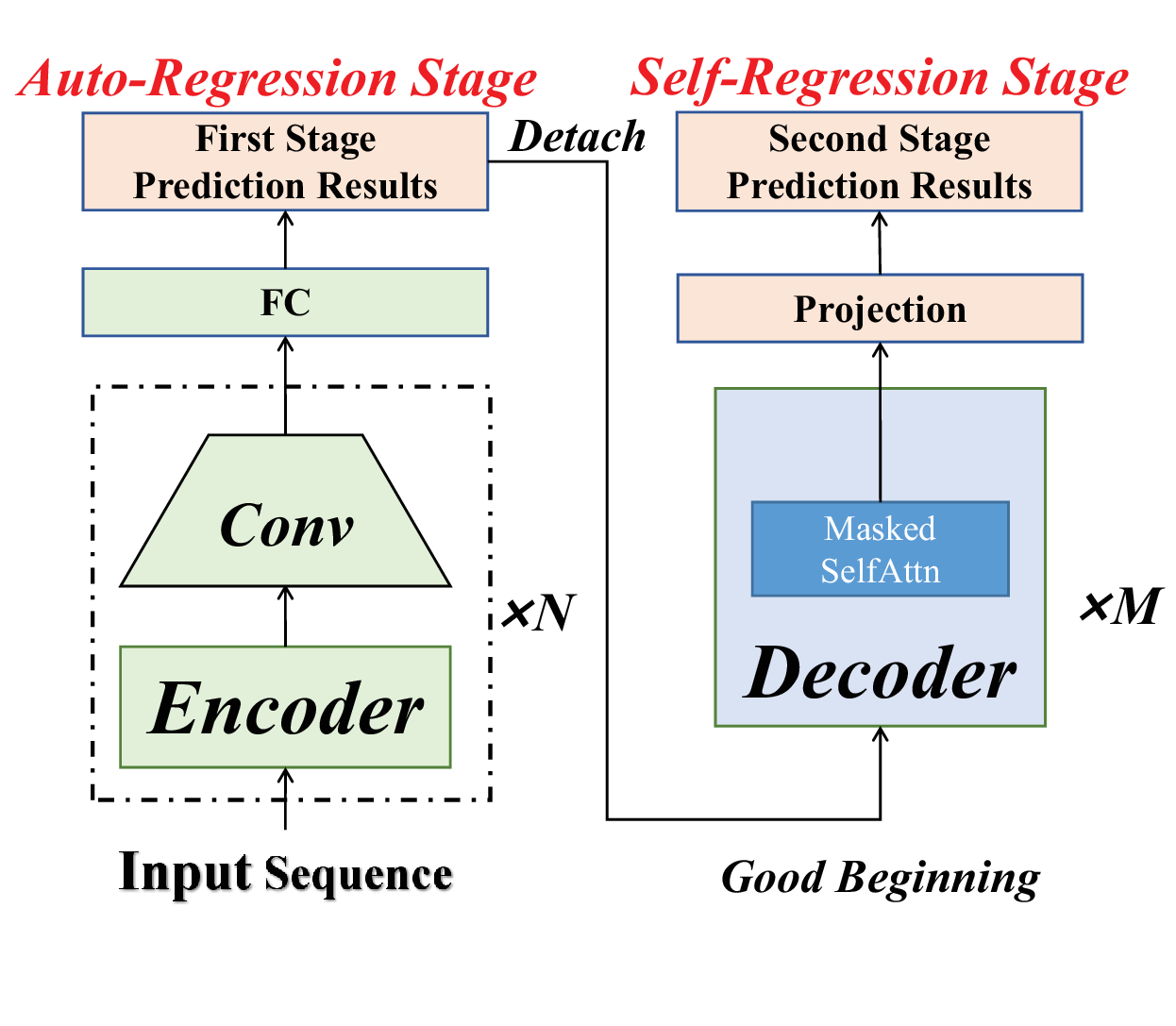}\label{fig2b}}	
	\caption{(a) is the architecture of canonical TSFT with start token and (b) is the architecture of GBT which decouples the forecasting process of TSFT into two stages.}
	\label{fig2}
\end{figure}
Thus, we propose \textbf{\emph{GBT}}, a novel two-stage \textbf{\emph{T}}ransformer framework with \textbf{\emph{G}}ood \textbf{\emph{B}}eginning. An overview of it is shown in Figure \ref{fig2b} while the canonical architecture of TSFT is shown in Figure \ref{fig2a} for comparison. The idea of GBT is simple: We divide the architecture of TSFT into two stages according to their different functions and outputs of the first stage are treated as inputs of the second stage. The first stage is the \textit{Auto-Regression} stage. Only encoders are employed to capture features of input sequences and a FC layer is employed to obtain prediction sequences. It aims to acquire \textit{better prediction elements initialization} for the second stage through features of input windows. As prediction results of the first stage are trained with the truth of prediction windows, these prediction results will have more similar statistical properties with those of truth compared with former initialization methods. Thus, they are `\textit{Good Beginning}' for the second stage input. Then the second stage is the \textit{Self-Regression} stage. As `\textit{Good Beginning}' is supposed to obtain auto-regressive results from input sequences, the second stage only needs to be concerned about relationships within prediction sequences. Therefore, it employs decoders which are only composed of masked self-attention modules and abandon modules including start token and cross-attention to reduce reduplicate computations. Networks of the second stage \textit{\textbf{do not participate}} in the training phase of the first stage and parameters in the first stage will be \textit{\textbf{fixed}} during the training phase of the second stage. It could be seen that compared with conventional TSFT framework, GBT not only alleviates the potential problem of different statistical properties of input/prediction windows when handling non-stationary time series by decoupled two-stage framework and `\textit{Good Beginning}' but also owns fewer parameters under the same conditions.\par
Though two-stage achitecture can give inputs of decoder, i.e., the second stage, a better initialization, deviations of first-stage prediction results from ground truths are different in different time stamps {\citep{LSTNet,direct}}. Hence, we employ {the} \textit{\textbf{E}}rror \textit{\textbf{S}}core \textit{\textbf{M}}odification module (ESM) in the second stage to utilize prediction results of the first stage in a more rational way.\par
Leveraging from these mechanisms, GBT tackles non-stationary time series and {the} accompanied over-fitting problem well. Combining with experiments results, our main contributions are summarized as follows:\par
\begin{enumerate}
	\item We propose \textbf{GBT}: a two-stage \textbf{\emph{T}}ransformer framework with \textit{\textbf{G}}ood \textbf{\textit{B}}eginning to alleviate {the} over-fitting problem caused by bad initialization methods of decoder inputs in TSFTs, {especially} when forecasting non-stationary time series.
	\item GBT decouples the forecasting process of TSFT into \textit{Auto-Regression} and \textit{Self-Regression} stages to solve the problem of different statistical properties of input/prediction windows when handling non-stationary time series. Specifically, the second \textit{Self-Regression} stage employs {the} \textbf{\textit{E}}rror \textbf{\textit{S}}core \textbf{\textit{M}}odification module (ESM) to utilize the \textit{Good Beginning} provided by \textit{Auto-Regression} stage in a more rational way.
	\item Experiments on seven benchmark datasets demonstrate that GBT outperforms SOTA TSFTs and other forecasting models with only canonical attention and convolution while owning less time and space complexity.
	\item The framework of GBT is general and adaptive enough to couple with other forecasting models, especially TSFTs, to achieve more promising performances.
\end{enumerate}

\section{Related Works}
\subsection{Time Series Forecasting}
Time series forecasting becomes a critical ingredient in various fields, such as stock prediction {\citep{FourierStock}}, traffic forecasting {\citep{Traffic1,Traffic2}}, sensor-based recognition {\citep{ahmad2019human}} and COVID-19 pandemic analysis {\citep{COVID-19,zhao2021empirical,COVID}}. Thus, various models have recently been proposed to forecast time series. With the development of deep learning and the need of tackling Long Sequence Time series Forecasting (LSTF) problem, traditional models, such as ARIMA {\citep{box1974,box2015}} and SSM {\citep{SSM}}, fail to challenge deep learning based models, including models based on RNN, CNN and Transformer.\par
\subsection{Deep Time Series Forecasting Methods}
RNN based methods {\citep{LSTM,LSTNet,deepar,WEERAKODY2021161}} predict time series through multi-step rolling procedures, which means that they could not capture long-term dependency well. CNN is another feasible approach {\citep{oord2016wavenet,scinet,TCN,Wibawa2022}}, which leverages dilated causal convolution to do time series forecasting. Different from the above two models, Transformer {\citep{attention2017}} shows better potential in capturing long-term dependency thanks to its self-attention mechanism, which helps model equally available to any part of time series sequences regardless of temporal distance. Recently, popular corresponding models or methods have already formed an approximately mature research field for time series forecasting Transformers (TSFT) {\citep{STtrans,informer,logtrans,Triformer,Pyraformer,Autoformer,FEDformer,ETSformer}}. After one-forward time series forecasting procedure was proposed by Informer {\citep{informer}}, one-forward TSFT becomes popular on account of its outstanding performance. The researches on enhancing performance of one-forward TSFT include solving its quadratic complexity problem {\citep{informer,logtrans,Triformer,Pyraformer}} and decoupling trend and seasonal components by Fourier transform {\citep{Autoformer,FEDformer,ETSformer}}. However, they all do not fundamentally modify the architecture of TSFT, which is different from this paper. That is also why GBT could be combined with them easily.\par
\subsection{Two-stage Forecasting Methods}
Some self-supervised time series forecasting methods based on representation learning {\citep{TS2Vec,Cost}} use two stage forecasting architectures. Their first stage is to seek the representation of input sequence through convolutions and obtain the universal feature map of input sequence through contrastive learning  {\citep{TS2Vec,Cost}}; then their second stage is able to acquire prediction results though representations outputted by the first stage and Regressors {\citep{TS2Vec,Cost}}.\par
{Additionally, some forecasting models are involved with diverse reconstruction strategies {\citep{N-HiTS, FiLM}}, which makes them analogous to `two-stage model'. Though most of them own only one training/inference phase, their networks can be divided into reconstruction and forecasting parts. The participation of the reconstruction part renders them able to be more interpretablely and precisely extracting feature maps of input sequences.\par
Obviously, either models with representation learning or with reconstruction have a chance to tackle distribution shifts of input sequences quite well. However, as we point out in Section \ref{introduction}, the distribution shift between input sequences and prediction sequences are more formidable to get around and it cannot be solved by these two techniques as they are only concerned with the features of input sequences.\par}
{The architecture of GBT is motivated from conventional two-stage models, but} usages of two-stage architecture of GBT proposed in this paper are different from {them} in three points: (1) Outputs of both two stages of GBT are prediction results rather than representations. (2) The intention of using two-stage architecture in GBT is to tackle non-stationarity instead of strengthening feature extraction capability of networks. (3) GBT is built upon Transformer while these self-supervised methods are mainly based on CNN/MLP.\par
\subsection{Non-stationarity Handling Technique}
Negative influences to time series forecasting originated from non-stationarity have already caught attentions in recent years {\citep{RevIN, non-stationaryTransformer}}. However, existing solutions are only involved of diverse normalization/de-normalization methods. RevIN {\citep{RevIN}} employs instance normalization and its reverse version at the beginning and ending of every single window. Non-stationary Transformer {\citep{non-stationaryTransformer}} applies similar methods within attention mechanism. It could be seen that they do not essentially change the framework of TSFT unlike GBT.\par
\section{Preliminary}
We provide the definition of time series forecasting problem and name origins of GBT's two stages.\par
\subsection{Time Series Forecasting Problem}
Given input window ${{\{z_{i,1:t_{0}}\}}_{i=1}^N}$, the task is to obtain prediction window ${{\{z_{i,t_{0}+1:T}\}}_{i=1}^N}$. $N$ is the number of variates. $t_{0}$ denotes the input window size and $T-t_{0}$ is the prediction window size. If ${N}=1$, we call it univariate time series forecasting. Otherwise, we call it multivariate time series forecasting.\par
One-forward/Direct {\citep{direct}} forecasting strategy, which predicts the whole prediction window simultaneously, becomes the mainstream forecasting formula in that it avoids possible error accumulation brought by recursive forecasting strategy {\citep{deepar, LSTNet}}. Nearly all forecasting methods discussed in this paper employ direct forecasting strategy.\par
{\subsection{Time Series Forecasting Transformer}
Time series forecasting Transformer (TSFT) is developed from the Vanilla Transformer {\citep{attention2017}}. Similar with the canonical one, time series forecasting Transformer has an encoder-decoder architecture. Here, we only introduce the architecture of TSFT with direct forecasting procedure and zero-intialization of decoder inputs, which is majorly analyzed in the main text. As shown in Figure \ref{figa1}, the encoder receives the value embedding of the past input sequence added with its position and time encoding features. Each layer of the encoder mainly contains a self-attention block and a feed-forward block. Decoder receives similar format of the following target sequence information but the unknown parts are padded into zero. Note that this is quite different from the canonical Transformer step-by-step `dynamic decoding' process, for decoder now could obtain the whole outputs through one forward procedure, which avoids potential error accumulation brought by recursive decoding. Decoder extra receives the feature map from the encoder for cross-attention. Each layer of the decoder mainly contains a masked self-attention block, a cross-attention block and a feed-forward block. Start token is a popular technique to be employed to give decoder inputs a better start {\citep{BERT,informer}}. However, it has been proved in our paper that it is not useful enough when tackling real-world non-stationary time series.\par
\begin{figure}
	\centering
	\begin{tabular}{c}
		\centering
		\includegraphics[width=10cm]{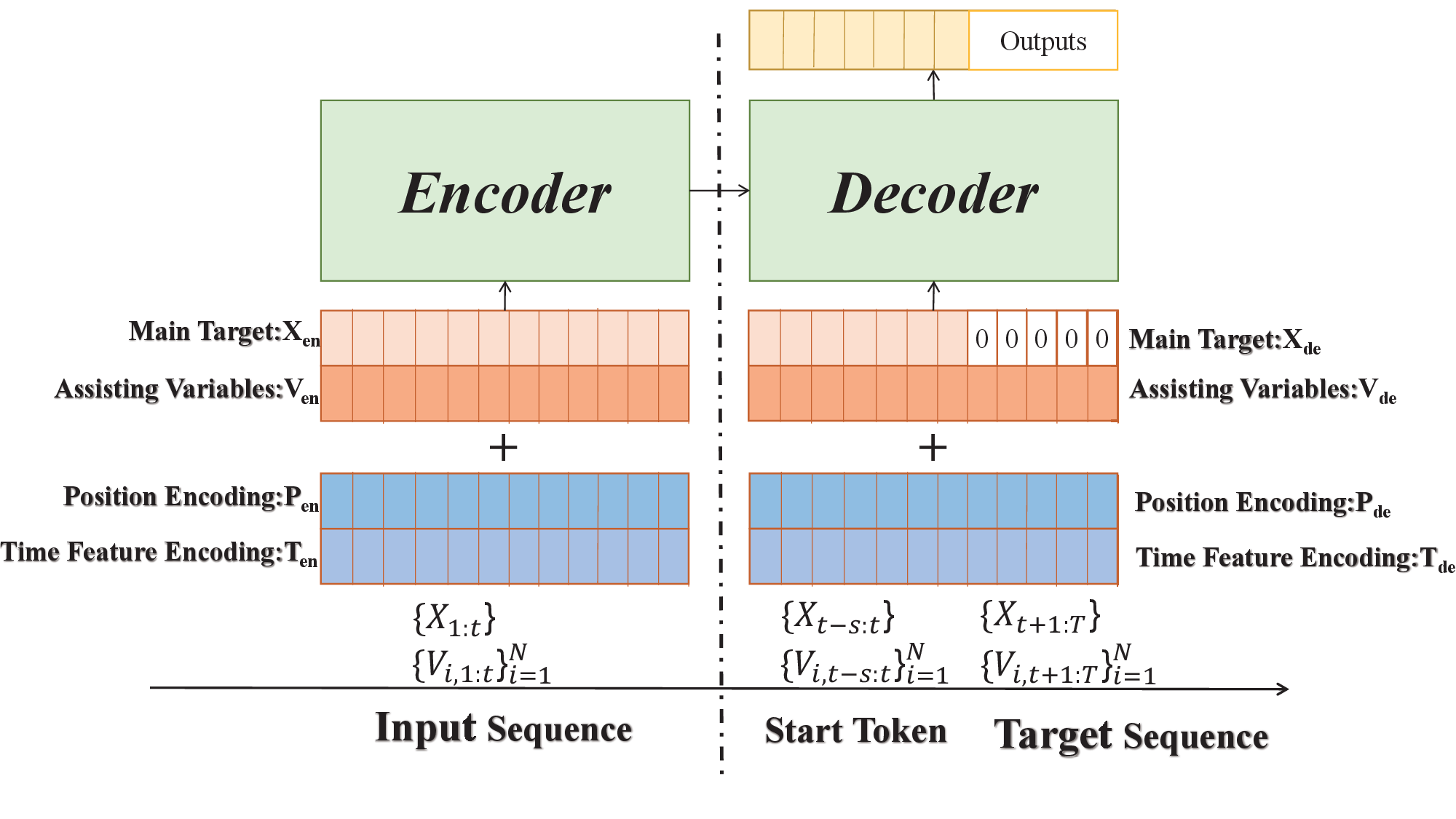}
	\end{tabular}
	\caption{Overview of the canonical architecture of time series forecasting Transformer. In this figure, we assume that the number of the main target to predict is one. $t$ is the input window size and $T-t$ is the prediction window size. $s$ stands for the length of the start token and $N$ refers to the number of the associating variables.}  
	\label{figa1}  
\end{figure}}
\subsection{Origin of Stage Names in GBT}
Canonical TSFTs are composed of encoders and decoders. The encoder and cross-attention module in decoder are \textit{Auto-Regression} processes which aim to seek connections between input and prediction windows. Meanwhile, masked self-attention module in decoder belongs to \textit{Self-Regression} process which is meant to deduce prediction elements from themselves. In fact, these two processes are both auto-regressive. We give them different names to distinguish them by their \textit{\textbf{different auto-regressive inputs}}. They also respectively correspond to two stages of GBT.\par
{\subsection{Meanings of Abbreviations and Phrases}
The meanings of mentioned abbreviations and phrases used in this paper are shown in Table \ref{taba1}.\par
\begin{table}[pos=h]
	\footnotesize
	\renewcommand{\arraystretch}{1.1}
	\centering
	\setlength\tabcolsep{10pt}
	\renewcommand{\multirowsetup}{\centering}
	\caption{Meanings of abbreviations and phrases}
	\label{taba1}	
	\begin{tabular}{cc}
		\toprule[1.5pt]
		Abbr./Phrase & Meaning  \\
		\midrule[1pt]
		TSFT        & \textbf{\textit{T}}ime \textbf{\textit{S}}eris \textbf{\textit{F}}orecasting \textbf{\textit{T}}ransformer  \\
		Auto-Regression& \makecell{Inference process of prediction elements by input sequences} \\
		Self-Regression & \makecell{Inference process of prediction elements by themselves}\\
		\bottomrule[1.5pt]
	\end{tabular}
\end{table}}
\section{Methodology}
\label{Methodology}
\subsection{Framework of GBT}
\subsubsection{Motivation}
\begin{figure*}[pos=b]
	\centering
	\includegraphics[width=0.8\columnwidth]{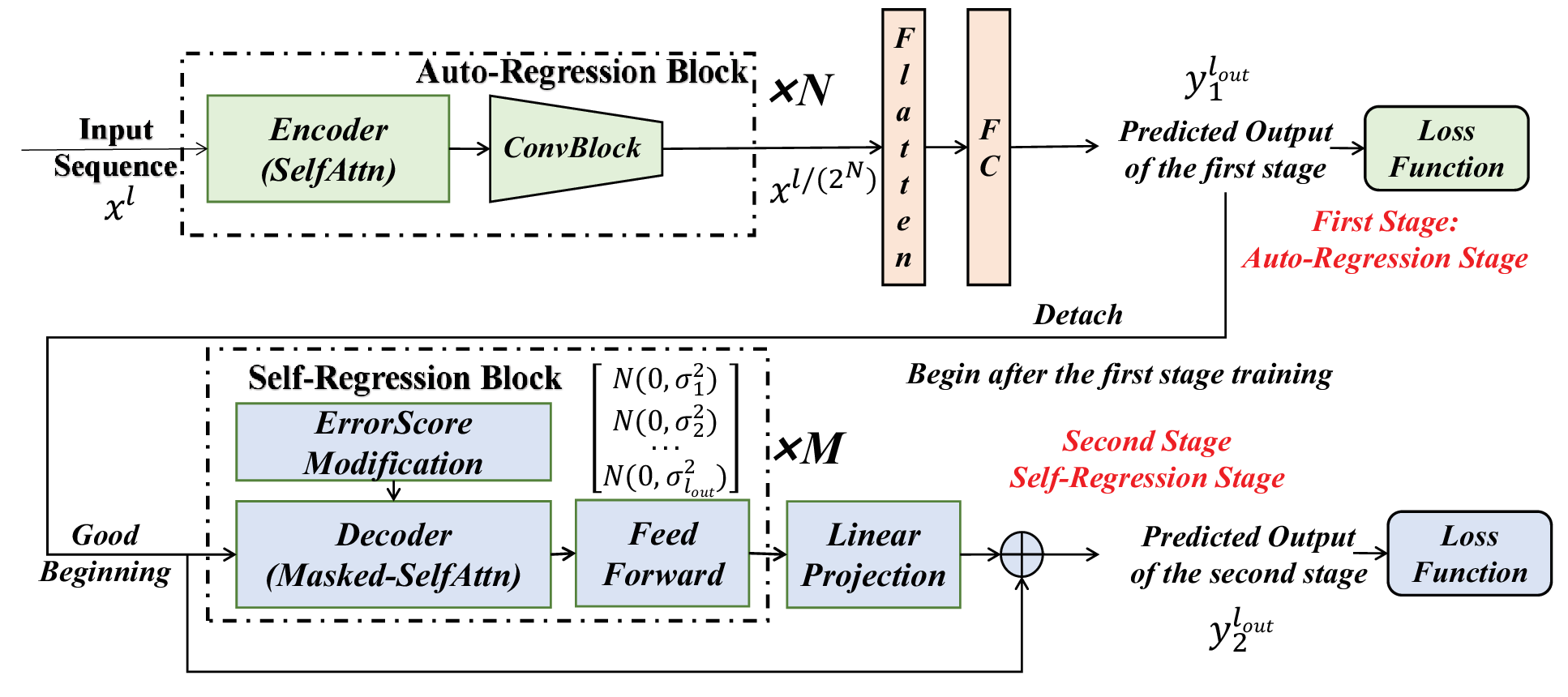}
	\caption{An overview of whole GBT. The upper half is the \textit{Auto-Regression} stage and the lower half is the \textit{Self-Regression} stage. These two parts are trained separately. The concrete architecture of ConvBlock is shown in Figure \ref{fig4a}. We omit the pyramid architecture in \textit{Auto-Regression} stage shown in Figure \ref{fig4b}.}
	\label{fig3}
\end{figure*}

We have briefly introduced the motivation of the GBT framework in Section \ref{introduction}, i.e., current two existing solutions to help initialize decoder inputs are not enough to handle {the} over-fitting problem of TSFTs caused by zero-initialized inputs in decoder and the non-stationarity of time series. We provide a more detailed analysis here to validate above statement.\par
The first solution is using start token in decoders to help initialize prediction elements {\citep{informer}}. This is the most common condition. Then let’s check the inference process of the first masked self-attention within the decoder. Its scaled dot-product could be expressed {as Equation \ref{eq_attn}}. $ \{Q_1/K_1, Q_2/K_2\} $ refer to query/key of \{start token, prediction elements\} and $d$ is the embedding dimension. Ignoring position embedding and bias, $ Q_2/K_2 $ will be zero tensors in that the linear projection of zero elements will also be zero elements. Then we can get query-key matching matrix as Equation \ref{eq1}. It could be seen that attention scores allocated to prediction elements are all zero so start token has no effect on inference process of prediction elements and could not solve aforementioned problem. \par
\begin{eqnarray}
	\label{eq_attn}	 {Attn(Q,K,V)=Softmax(\frac{QK^\top}{\sqrt{d}})V,\ \ Q=[Q_1,Q_2]^\top,\ \ K=[K_1,K_2]^\top }
\end{eqnarray}

\begin{eqnarray}
	\label{eq1}
	\begin{split}
		QK^\top &=
		\begin{bmatrix}
			Q_1 \\
			Q_2  
		\end{bmatrix}\bigg|_{Q_2=0}
		\times 
		[K_1^\top, K_2^\top]|_{K_2=0} 
		= \begin{bmatrix}
			Q_1K_1^\top & 0 \\
			0 & 0 
		\end{bmatrix}
	\end{split}
\end{eqnarray}

Some TSFTs additionally employ decomposed trend terms of input sequences from encoder {\citep{Autoformer, FEDformer}} to initialize inputs of decoders. However, currently $ Q_1,K_1,Q_2,K_2 $ are generated from components of input sequences in Equation \ref{eq1}. This is a better initialization for TSFTs when dealing with stationary time series. However, when facing non-stationary time series, it is no longer effective as trends of input/prediction sequence may be different. Therefore, we propose two-stage GBT, which decouples prediction processes of TSFT and transforms them into two different stages, i.e., the first \textit{Auto-Regression} stage and the second \textit{Self-Regression} stage, to solve this problem better. Its architecture is shown in Figure \ref{fig3}.\par
\subsubsection{Auto-Regression Stage}
\begin{figure}[t]
	\centering
	\subfloat[]{\includegraphics[width=0.38\columnwidth]{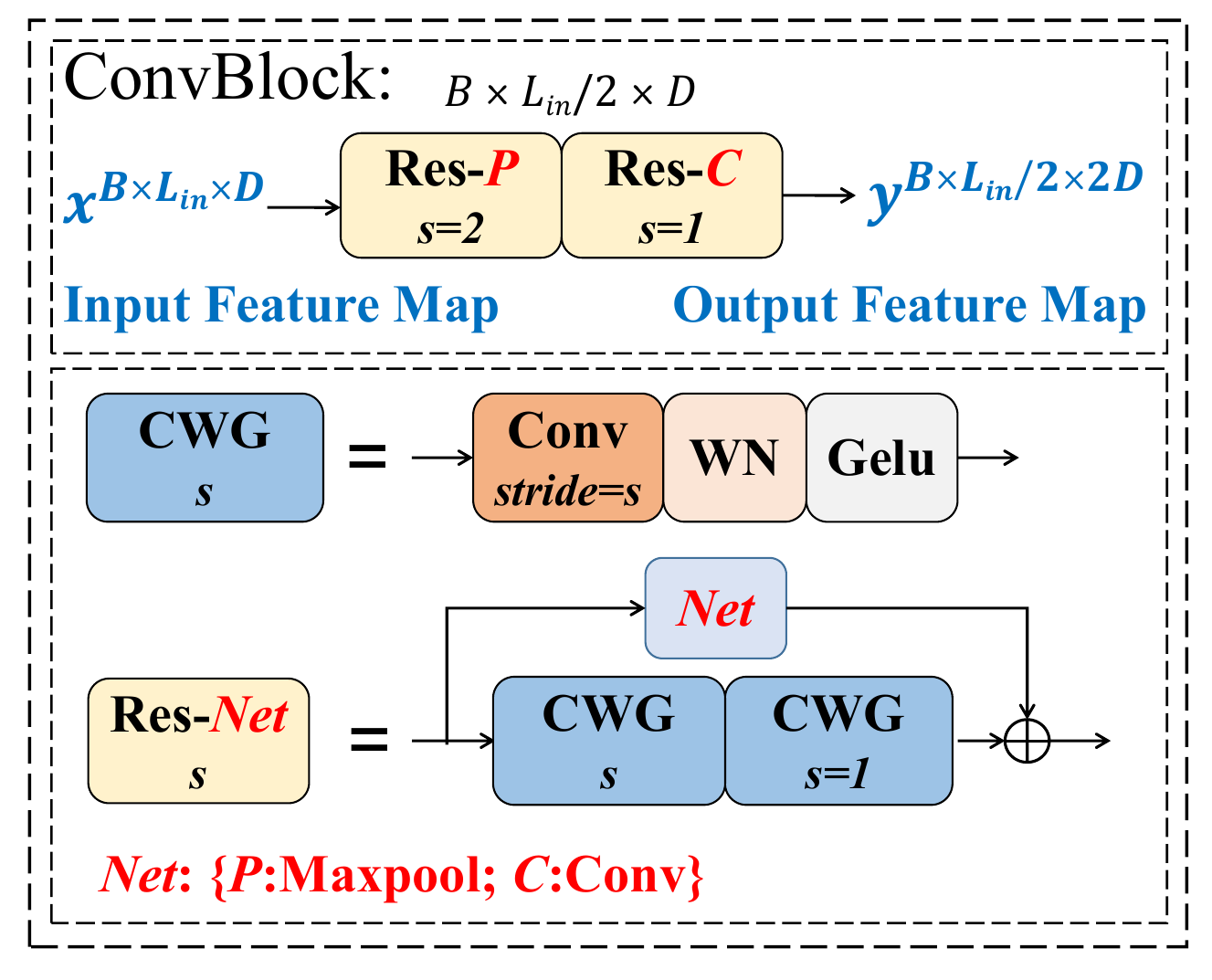} \label{fig4a}}
	\subfloat[]{\includegraphics[width=0.33\columnwidth]{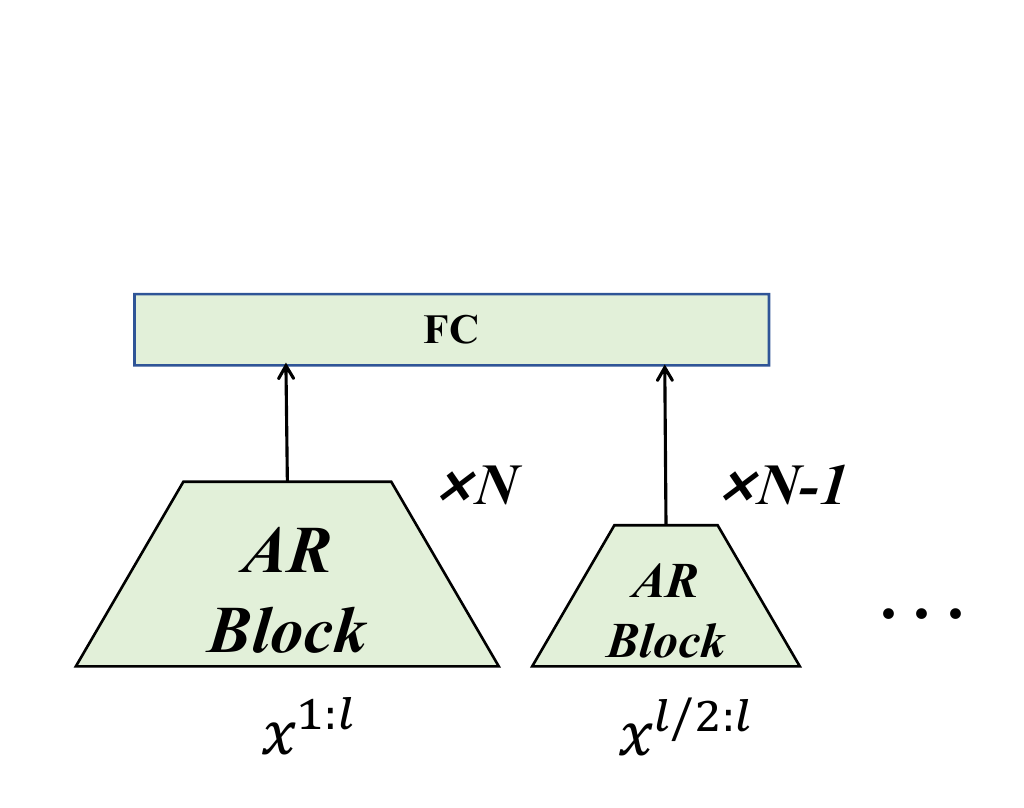} \label{fig4b}}
	\caption{ConvBlock shown in (a) is composed of Res-$P$ and Res-$C$. Res-$P$ is employed to shrink the sequence length by half and Res-$C$ is employed to double the hidden dimension. (b) is the pyramid architecture similar with Informer. AR Block refers to the \textit{Auto-Regression} Block. Extra pyramid networks own fewer AR Blocks with shorter input sequence.}
	\label{fig4}
\end{figure}

The first \textit{Auto-Regression} stage is composed of $ N $ \textit{Auto-Regression} Blocks to extract feature maps of input sequences. As Figure \ref{fig3} shows, each \textit{Auto-Regression} Block consists of one encoder and one ConvBlock. Its encoder is the same as the encoder of canonical Transformer but removing the feed-forward layer. Motivated by \citet{ResNet,RTNet}, ConvBlock replaces the feed-forward layer and uses convolutions to shrink the sequence length and double the hidden dimension shown in Figure \ref{fig4a}. In each ConvBlock, WN {\citep{WN}} is chosen as the normalization method and Gelu {\citep{GELU}} is employed as the activation function. ConvBlock is meant to enhance the locality of attention module {\citep{Reformer}} and smooth the turbulence {\citep{ETSformer}} brought by potential anomalies. We also employ similar pyramid network like Informer {\citep{informer}} to extract hierarchical feature maps as shown in Figure \ref{fig4b}. Finally, we employ a FC layer to obtain prediction results at the first stage and to replace the cross-attention module of canonical TSFT which has the similar function of obtaining prediction sequence (feature map) through input sequence feature map in decoders. This replacement is established and rational in two reasons: (1) They share similar functions, i.e., \textit{Auto-Regression} of prediction sequences through input sequences; (2) Currently the first stage only receives input sequences so that cross-attention module is unable to be applied. Consequently, prediction elements of the first stage are trained through the loss function and ground truths of prediction windows, so they will have more similar statistical properties with those of truth compared with former initialization methods. Thus, they are `\textit{Good Beginning}' for the second stage input. Note that networks of the second stage \textit{do not participate} in the training of the first stage.\par
\subsubsection{Self-Regression Stage}
After the training of the first stage, we have acquired \textit{Auto-Regression} parts of prediction elements through input sequences and it is `\textit{Good Beginning}' for the input of the second \textit{Self-Regression} stage. For the second stage is an inference process of prediction elements by themselves, it contains $ M $ decoders composed of a masked self-attention module and a corresponding feed-forward layer. Compared with the decoder of canonical Transformer, it abandons cross-attention in that it is an \textit{Auto-Regression} module which is redundant in \textit{Self-Regression} process. Finally, we use a linear projection layer to obtain prediction results. Parameters of the first stage network will be \textit{fixed} during the training phase of the second stage. Additionally, within the second stage, we employ {the} Error Score Modification module to manage the error accumulation of `\textit{Good Beginning}' provided by the first stage, which is shown as following.\par
\subsection{Error Score Modification}

{\subsubsection{Motivation}}
As `\textit{Good Beginning}' provided by the first stage is the result of \textit{Auto-Regression}, it can be deduced from former analysis in Section \ref{introduction} and Figure \ref{fig1b} that later prediction elements will be further from truth in the statistical sense. {This phenomenon is elaborated in \citep{direct}. Therefore, we need an additional mechanism specified for this problem to rationally deduce the prediction sequence from the auto-regressive results of the first stage. As the later prediction elements of the output sequence of the first stage is less reliable, it is reasonable to guide the second stage to allocate more scores to earlier elements, which are more conceivable to approach the real value, during the attention.}\par
{\subsubsection{Architecture}}
We adopt a group of learnable Gauss distribution $ N(0, \sigma_i^2), i\in[1,l_{out}] $ {\citep{AnomalyTransformer}} and add it into the masked query-key matching matrix in the second stage shown in Figure \ref{fig3}. Its concrete formula is presented in Equation \ref{eq4}. $ QK^\top $ is the query-key matching matrix with size $ l_{out}\times l_{out} $ where $ l_{out} $ is the prediction sequence length. $ N(0,\sigma_i^2) $ is a learnable Gauss distribution where the center is zero and the scale $ \sigma $ is a learnable parameter. The center is fixed to zero in ESM so that more masked self-attention scores will be allocated to earlier and more trusted prediction elements, which means that earlier prediction elements will make more contributions for the \textit{Self-Regression} of any prediction element during the second stage. The $ \sigma_i $ is learnable so that ESM will be more adaptive. The pseudo-code of ESM is given as Algorithm \ref{alg1}.\par
\begin{eqnarray}
	\label{eq4}
	ESM(Q,K) = QK^\top + \begin{bmatrix}
		N(0,\sigma_1^2)\\ 
		N(0,\sigma_2^2)\\ 
		\cdots \\
		N(0,\sigma_{l_{out}}^2)\\
	\end{bmatrix}
\end{eqnarray}
\begin{algorithm}
	\caption{Masked self-attention with ESM} 
	\label{alg1} 
	\renewcommand{\algorithmicrequire}{\textbf{Input:}}
	\renewcommand{\algorithmicensure}{\textbf{Output:}}
	\begin{algorithmic}[1]
		\REQUIRE Tensor $ x\in \mathbb{R}^{L\times d}$
		\renewcommand{\algorithmicrequire}{\textbf{Layer params:}}
		\REQUIRE $ Linear(x) $: Linear projection layer of $ x $
		\STATE $Q,K,V = Linear _{1} (x)$    $(Q,K,V\in \mathbb{R}^{L\times d}) $
		\STATE $\sigma$ = $Linear_{2}(x)$  $ ( \sigma\in \mathbb{R}^{L\times 1} ) $
		\STATE Calculate self-attention scores $A= QK^\top$ ($A\in\mathbb{R}^{L\times L}$)
		\STATE $ \sigma $= Sigmoid($5\sigma$) + 1e-5
		\STATE $\sigma = 3^\sigma - 1$
		\STATE $\sigma$ = Repeat($\sigma$, dim=1) ($ \sigma\in\mathbb{R}^{L\times L} $)
		\STATE Generate Gauss distribution matrix $G{\sim}			N(0,\sigma^2)$   ($ G\in\mathbb{R}^{L\times L} $)
		\STATE Add self-attention scores with ESM: $A = A + G$
		\STATE  $A = mask(A) $
		\ENSURE  Masked self-attention scores $ A $
	\end{algorithmic}
\end{algorithm}
\subsection{Complexity Analysis}
Essentially, GBT is a transformed and decoupled Transformer framework with less time and space complexity: (1) Feed-forward layers in encoders are replaced with ConvBlocks which shrink input sequence length, thus mediately decreasing complexity of self-attention; (2) Due to two-stage architecture, start token is no longer needed so the sequence length within decoder of GBT is shorter than that of canonical TSFT; (3) A FC layer is employed to play the role of cross-attention in the \textit{Auto-Regression} stage. Though these two modules both have $O(L ^{2} )$ complexity, GBT gets rid of plenty of linear projection layers in cross attention modules which bring considerable computation. Note that GBT is only a framework so any modified TSFT could be introduced into GBT to acquire less complexity. We validate above analysis in later Section \ref{Complexity}.\par
\section{Experiment}
\subsection{Dataset}
To evaluate the forecasting capability of our proposed GBT, we perform experiments on real-world datasets. Currently, there are dozens of benchmark time series forecasting datasets which are widely chosen by other state-of-the-art forecasting models {\citep{Autoformer, FEDformer, scinet, TS2Vec}}. Among them, we choose \{ETTh$ _{1} $, ETTm$ _{2} $\} in \{ETTh$ _{1} $, ETTh$ _{2} $, ETTm$ _{1} $, ETTm$ _{2} $\} in that these two have different sampling intervals and are more commonly used. The rest of chosen datasets are \{ECL, WTH, weather, Traffic, Exchange, ILI\}. We briefly introduce these eight selected datasets\footnote{\{ETTh$_1$, ETTm$_2$, ECL, weather, Traffic, Exchange, ILI\} datasets were acquired at: \url{https://drive.google.com/drive/folders/1ZOYpTUa82_jCcxIdTmyr0LXQfvaM9vIy?usp=sharing}}\footnote{WTH dataset was acquired at: \url{https://www.ncdc.noaa.gov/orders/qclcd/}} and their usages in this paper below and their numeral details (Size, Dimension and Frequency) are shown in Table \ref{taba2}:\par
\begin{table}[]   
	\caption{Details of eight datasets}
	\footnotesize
	\label{taba2}
	\renewcommand{\arraystretch}{1.1}
	\setlength\tabcolsep{10pt}
	\begin{tabular}{cccc}
		\toprule[1.5pt]
		Dataset  & Size  & Dimension & Frequency \\
		\midrule[1pt]
		ETTh$ _{1} $    & 17420 & 7 & 1h\\
		ETTm$ _{2} $    & 69680 & 7 & 15min    \\	
		WTH&35064&7&1h\\
		ECL& 26304 & 321& 1h\\
		Traffic  & 17544 & 862& 1h\\	
		Exchange & 7588  & 8 & 1day    \\			
		weather &52696 &21 &10min	\\
		ILI&966& 8& 7days\\
		\bottomrule[1.5pt]
	\end{tabular}
\end{table}	

\paragraph{ETT}
(Electricity Transformer Temperature) dataset {\citep{informer}} is composed of ETT data collected from two separated counties in China lasting for almost 2 years. It contains four subsets: \{ETTh$_1$, ETTh$_2$\} are 1-hour-level datasets; \{ETTm$_1$, ETTm$_2$\} are 15-min-level datasets. Each data point is composed of the target value `OT' (oil temperature) and other 6 power load features. For averting unnecessary experiments, we choose one 1-hour-level subset, ETTh$_1$, and one 15-min-level subset, ETTm$_2$, to do experiments on. The train/val/test is 12/4/4 months.\par 
\paragraph{WTH}
dataset contains local climatological data from nearly 1,600 U.S. locations. It is a 1-hour-level dataset spanning 4 years from 2010 to 2013. Each data point consists of the target value `WetBulbCelsius' and 11 climate features. The train/val/test is 60\%/20\%/20\% according to settings of other baselines {\citep{informer, Cost}}.\par
\begin{table}[pos=b]
	\footnotesize      
	\renewcommand{\arraystretch}{1.1}
	\centering
	\setlength\tabcolsep{2pt}
	\renewcommand{\multirowsetup}{\centering}
	\caption{Details of hyper-parameters/settings}
	\label{taba3}	
	\begin{tabular}{cc}
		\toprule[1.5pt]
		Hyper-Parameters / Settings&Values / Mechanisms       \\
		\midrule[1pt]
		The number of pyramid networks&3\\
		\makecell{The number of AR Blocks in the main pyramid network} &3 \\
		\makecell{Embedding dimensions of \textit{Auto-Regression} stage}&32      \\
		\makecell{Embedding dimensions of \textit{Self-Regression} stage}&512     \\
		The kernel size of Conv layers&3   \\
		Input window size &96 (36 only for ILI)    \\
		Standardization&Z-score         \\
		Loss function&MSE     \\
		
		Optimizer&Adam\\
		Dropout&0.1     \\
		Learning rate&1e-4\\
		Learning rate decreasing rate&Half per epoch\\
		
		Batch size&32 \\
		Random seed&4321 (if used)     \\
		Platform&\makecell{Python 3.8.0; Pytorch 1.11.0 }       \\
		Device&\makecell{A single NVIDIA GeForce RTX 3090 24GB GPU} \\
		\bottomrule[1.5pt]    
	\end{tabular}
\end{table}

\paragraph{ECL}
(Electricity Consuming Load) dataset contains time series of electricity consumption (Kwh) from 321 clients. It is a 1-hour-level dataset which is converted into 2 years by {\citet{informer}} and `MT\_321' {\citep{ETSformer, FEDformer, Autoformer}} is the target value according to the most common settings  (some papers use `MT\_001' {\citep{TS2Vec}}/`MT\_320'{\citep{informer}}, but they are not the most commonly used). The train/val/test is 70\%/10\%/20\% following \citet{ETSformer, FEDformer, Autoformer}.
\paragraph{Traffic}
dataset is collected hourly from the California Department of Transportation to describe road occupancy rates in San Francisco Bay area freeways spanning 48 months (2015-2016). The train/val/test is 70\%/10\%/20\%. We set the variate name of target as `OT' in \{Traffic, Exchange, weather, ILI\} for convenience according to \citet{FEDformer, LSTNet}.
\begin{table}[pos=b]
	\footnotesize   
	\renewcommand{\arraystretch}{1.1}
	\centering
	\setlength\tabcolsep{4pt}
	\renewcommand{\multirowsetup}{\centering}
	\caption{Results of multivariate forecasting}
	\label{tab1}
	\begin{tabular}{c|c|cccc|cccc|cccc}
		\toprule[1.5pt]
		\multirow{2}{*}{Methods}     & \multirow{2}{*}{Metrics} & \multicolumn{4}{c|}{ETTh$ _{1} $}       & \multicolumn{4}{c|}{ETTm$ _{2} $}       & \multicolumn{4}{c}{ECL}\\
		\cmidrule(lr){3-6}
		\cmidrule(lr){7-10}
		\cmidrule(lr){11-14}
		 && 96& 192& 336& 720& 96& 192& 336& 720& 96& 192& 336& 720\\
		\midrule[1pt]
		\multirow{2}{*}{GBT-Vanilla} & MSE & \textit{\textbf{0.398}} & \textit{\textbf{0.448}} & \textit{\textbf{0.497}} & \textit{\textbf{0.538}} & \textit{\textbf{0.189}} & \textit{\textbf{0.249}} & \textit{\underline{0.324}}& \textit{\textbf{0.395}} & \textit{\textbf{0.143}} & \textit{\textbf{0.175}} & \textit{\textbf{0.197}} & \textit{\underline{0.235}}\\
		& MAE & \textit{\textbf{0.418}} & \textit{\textbf{0.442}} & \textit{\textbf{0.470}} & \textit{\textbf{0.505}} & \textit{\textbf{0.276}} & \textit{\underline{0.324}}& 0.368& 0.419& \textit{\textbf{0.246}} & \textit{\textbf{0.277}} & \textit{\textbf{0.298}} & \textit{\textbf{0.336}} \\
		\multirow{2}{*}{FEDformer}  & {MSE} & {0.415} & {0.474} & {0.535} & {0.680} & {0.203} & {0.269} & {0.325}          & {0.421}          & {0.193} & {\textit{\underline{0.201}}} & {0.214} & {0.246} \\
		& {MAE} & {0.453} & {0.493} & {0.524} & {0.593} & {0.287} & {0.328} & {\textit{\underline{0.366}}} & {\textit{\underline{0.415}}} & {0.308} & {0.315}          & {0.329} & {0.355}
		\\
		\multirow{2}{*}{Pyraformer}  & MSE & 0.662& 0.791& 0.902& 0.974& 0.378& 1.192& 1.176& 6.720& 0.418& 0.408& 0.410& 0.407\\
		& MAE & 0.611& 0.683& 0.734& 0.780& 0.456& 0.870& 1.033& 2.077& 0.460& 0.454& 0.457& 0.456\\
		\multirow{2}{*}{ETSformer}   & MSE & 0.511& 0.561& 0.599& 0.588& \textit{\underline{0.189}}& \textit{\underline{0.253}}& \textit{\textbf{0.314}} & \textit{\underline{0.414}}& \textit{\underline{0.187}}& 0.199& \textit{\underline{0.212}}& \textit{\textbf{0.233}} \\
		& MAE & 0.487& 0.513& 0.529& \textit{\underline{0.541}}& 0.280& \textit{\textbf{0.319}} & \textit{\textbf{0.357}} & \textit{\textbf{0.413}} & 0.304& 0.315& 0.329& 0.345\\
		& {MSE} & {1.061} & {1.108} & {1.130} & {1.179} & {0.812} & {0.884} & {1.263} & {4.174} & {0.382} & {0.381} & {0.383} & {0.390} \\
		\multirow{-2}{*}{{AirFormer}} & {MAE} & {0.813} & {0.836} & {0.846} & {0.886} & {0.734} & {0.766} & {0.873} & {1.660} & {0.441} & {0.443} & {0.445} & {0.452}\\
		\multirow{2}{*}{SCINet}      & MSE & 0.531& 0.535& 0.584& 0.685& 0.312& 0.573& 1.870& 3.462& 0.210& 0.234& 0.227& 0.269\\
		& MAE & 0.503& 0.513& 0.560& 0.634& 0.415& 0.591& 1.078& 1.753& 0.333& 0.345& 0.340& 0.373\\
		\multirow{2}{*}{TS2Vec}      & MSE & 0.670& 0.781& 0.911& 1.059& 0.360& 0.534& 0.833& 1.906& 0.336& 0.337& 0.350& 0.375\\
		& MAE & 0.588& 0.651& 0.718& 0.794& 0.426& 0.537& 0.694& 1.054& 0.412& 0.415& 0.426& 0.438\\
		\multirow{2}{*}{DLinear}    & {MSE} & {0.431} & {0.474} & {0.518} & {\textit{\underline{0.560}}} & {0.199} & {0.287} & {0.387} & {0.544} & {0.246} & {0.246} & {0.260} & {0.294} \\
		& {MAE} & {0.451} & {0.479} & {0.508} & {0.559}          & {0.395} & {0.363} & {0.429} & {0.506} & {0.345} & {0.348} & {0.361} & {0.388}
		\\
		\multirow{2}{*}{N-HiTS}      & MSE & \textit{\underline{0.411}}& \textit{\underline{0.465}}& \textit{\underline{0.523}}& 0.592& 0.192& 0.284& 0.346& 0.520& 0.198& 0.205& 0.218& 0.254\\
		& MAE & \textit{\underline{0.419}}& \textit{\underline{0.453}}& \textit{\underline{0.492}}& 0.556& \textit{\underline{0.277}}& 0.350& 0.376& 0.485& \underline{\textit{0.291}}& \textit{\underline{0.301}}& \textit{\underline{0.314}}& \textit{\underline{0.342}}\\
		\multirow{2}{*}{FiLM} & MSE & 0.700 & 0.718 & 0.721 & 0.722 & 0.230 & 0.285 & 0.339 & 0.433 & 0.845 & 0.849 & 0.861 & 0.891 \\
		& MAE & 0.555 & 0.570 & 0.579 & 0.604 & 0.307 & 0.338 & 0.370 & 0.420 & 0.761 & 0.761 & 0.764 & 0.774\\
		\midrule
		\midrule
		\multirow{2}{*}{Methods}     & \multirow{2}{*}{Metrics} & \multicolumn{4}{c|}{WTH}& \multicolumn{4}{c|}{Traffic}     & \multicolumn{4}{c}{Exchange}    \\
		\cmidrule(lr){3-6}
		\cmidrule(lr){7-10}
		\cmidrule(lr){11-14}
		&   & 96& 192& 336& 720& 96& 192& 336& 720& 96& 192& 336& 720\\
		\midrule[1pt]
		\multirow{2}{*}{GBT-Vanilla} & MSE & \textit{\textbf{0.434}} & \textit{\textbf{0.481}} & \textit{\textbf{0.514}} & \textit{\textbf{0.523}} & \textbf{\textit{0.509}}& \textbf{\textit{0.520}}& \textbf{\textit{0.535}}& \textbf{\textit{0.575}}& 0.110& \textit{\textbf{0.179}} & {{0.358}}& \textit{\textbf{0.756}} \\
		& MAE & \textit{\textbf{0.466}} & \textit{\textbf{0.506}} & \textit{\textbf{0.527}} & \textit{\textbf{0.532}} & \textbf{\textit{0.282}}& \textbf{\textit{0.293}}& \textbf{\textit{0.307}}& \textbf{\textit{0.317}}& 0.249& \textit{\underline{0.312}}& {{0.446}}& \textit{\textbf{0.655}} \\
		\multirow{2}{*}{FEDformer}  & {MSE} & {0.509} & {0.581} & {0.630} & {0.580} & {0.587} & {0.604} & {0.621} & {0.626} & {0.148} & {0.271} & {0.460} & {1.195} \\
		& {MAE} & {0.513} & {0.557} & {0.636} & {0.586} & {\textit{\underline{0.366}}} & {0.373} & {0.383} & {0.382} & {0.278} & {0.380} & {0.500} & {0.841}
		\\
		\multirow{2}{*}{Pyraformer}  & MSE & 0.540& 0.575& 0.593& 0.623& 0.938& 0.939& 0.948& ---   & 1.489& 1.642& 1.744& 2.080\\
		& MAE & 0.546& 0.567& 0.578& 0.599& 0.490& 0.488& 0.488& ---   & 1.018& 1.075& 1.107& 1.197\\
		\multirow{2}{*}{ETSformer}   & MSE & 0.538& 0.615& 0.655& 0.719& 0.607& 0.621& 0.622& 0.632& \textit{\textbf{0.085}} & \textit{\underline{0.182}}& \textit{\textbf{0.348}} & 1.025\\
		& MAE & 0.521& 0.566& 0.589& 0.624& 0.392& 0.399& 0.396& 0.396& \textit{\textbf{0.204}} & \textit{\textbf{0.303}} & \textit{\textbf{0.428}} & 0.774\\
		{}                            & {MSE} & {0.504} & {0.562} & {0.580} & {0.619} & {0.849} & {0.856} & {0.866} & {1.030} & {0.938} & {1.000} & {1.164} & {1.720} \\
		\multirow{-2}{*}{{AirFormer}} & {MAE} & {0.520} & {0.560} & {0.570} & {0.595} & {0.476} & {0.478} & {0.480} & {0.613} & {0.840} & {0.870} & {0.932} & {1.115}\\
		\multirow{2}{*}{SCINet}      & MSE & 0.489& 0.526& 0.572& 0.617& 0.581& 0.595& ---   & ---   & 0.221& 0.323& 0.661& 2.691\\
		& MAE & 0.495& 0.524& 0.562& 0.586& 0.423& 0.429& ---   & ---   & 0.365& 0.442& 0.564& 1.320\\
		\multirow{2}{*}{TS2Vec}      & MSE & \textit{\underline{0.450}}&\textit{\underline{0.505}}& \textit{\underline{0.532}}& \textit{\underline{0.566}}& 0.941& ---   & ---   & ---   & 0.184& 0.373& 0.666& 2.941\\
		& MAE & \textit{\underline{0.472}}& \textit{\underline{0.515}}& \textit{\underline{0.533}}& \textit{\underline{0.557}}& 0.550& ---   & ---   & ---   & 0.315& 0.452& 0.612& 1.313\\
		\multirow{2}{*}{DLinear}     & {MSE} & {0.539} & {0.592} & {0.610} & {0.653} & {0.725} & {0.665} & {0.674} & {0.716} & {0.219} & {0.350} & {0.563} & {1.076} \\
		& {MAE} & {0.522} & {0.557} & {0.571} & {0.600} & {0.460} & {0.438} & {0.441} & {0.457} & {0.387} & {0.481} & {0.606} & {0.799}
		\\
		\multirow{2}{*}{N-HiTS}      & MSE & 0.488& 0.539& 0.565& 0.628& \textit{\underline{0.561}}&\textit{\underline{0.543}}&\textit{\underline{0.554}}&\textit{\underline{0.616}} & \textit{\underline{0.099}}& 0.297& 0.576& 1.288\\
		& MAE & 0.496& 0.536& 0.555& 0.596& 0.369&\textit{\underline{0.352}}&\textit{\underline{0.357}}&\textit{\underline{0.377}}  & \textit{\underline{0.225}}& 0.390& 0.550& 0.847  \\
		\multirow{2}{*}{FiLM} &  MSE & 0.515 & 0.585 & 0.619 & 0.688 & 1.409 & 1.412 & 1.428 & 1.451 & 0.141 & 0.216 & \textit{\underline{0.351}} & \textit{\underline{0.938}} \\
		& MAE & 0.497 & 0.540 & 0.563 & 0.602 & 0.799 & 0.802 & 0.806 & 0.809 & 0.272 & 0.342 & \textit{\underline{0.434}}& \textit{\underline{0.736}}\\
		\bottomrule[1.5pt]   
	\end{tabular}
\end{table}

\paragraph{Exchange}
dataset {\citep{LSTNet}} is a 1-day-level dataset, including daily exchange rates of eight countries from 1990 to 2016. The train/val/test is 70\%/10\%/20\%.
\paragraph{weather}
dataset is a 10-min-level dataset which contains 21 meteorological indicators in Germany during 2020. The train/val/test is 70\%/10\%/20\%.\par
\paragraph{ILI} (Influenza-like Illness) dataset, is a 1-week-level dataset containing recorded influenza-like illness patients data in USA between 2002 and 2021. The train/val/test is 70\%/10\%/20\%.\par
\subsection{Baselines}
As our proposed GBT aims to alleviate {the} over-fitting problem of TSFTs when facing non-stationary time series, we select {six} state-of-the-art TSFTs, i.e., \{Informer {\citep{informer}}, Autoformer {\citep{Autoformer}}, FEDformer {\citep{FEDformer}}, Pyraformer {\citep{Pyraformer}}, ETSformer {\citep{ETSformer}}, {Airformer \citep{AirFormer}}\}, as baselines. We additionally choose other {six} state-of-the-art forecasting models not based on Transformer, i.e., SCINet {\citep{scinet}}, TS2Vec {\citep{TS2Vec}}, {FiLM \citep{FiLM}} N-BEATS {\citep{NBEATS}}, N-HiTS {\citep{N-HiTS}} and DLinear {\citep{DLinear}}, as baselines for further comparison. {Specially, Airformer, TS2Vec, FiLM and N-HiTS are involved with reconstruction or representation learning strategies.} Brief introductions of these {twelve} baselines are shown in Appendix \ref{baselines}.\par
\subsection{Experiment Details}
\begin{table}[pos=b]
	\footnotesize 
	\renewcommand{\arraystretch}{1.1}
	\centering
	\setlength\tabcolsep{4pt}
	\renewcommand{\multirowsetup}{\centering}
	\caption{Results of univariate forecasting}
	\label{tab2}
	\begin{tabular}{c|c|cccc|cccc|cccc}
		\toprule[1.5pt]
		\multirow{2}{*}{Methods}     & \multirow{2}{*}{Metrics} & \multicolumn{4}{c|}{ETTh$ _{1} $} & \multicolumn{4}{c|}{ETTm$ _{2} $} & \multicolumn{4}{c}{ECL}   \\
		\cmidrule(lr){3-6}
		\cmidrule(lr){7-10}
		\cmidrule(lr){11-14}
		&   & 96& 192& 336& 720& 96& 192& 336& 720& 96& 192& 336& 720\\
		\midrule[1pt]
		\multirow{2}{*}{GBT-Vanilla} & MSE& \textit{\textbf{0.051}} & \textit{\textbf{0.074}} & \textit{\textbf{0.080}} & \textit{\underline{0.119}} & \textit{\textbf{0.068}} & \textit{\textbf{0.091}} & \textit{\textbf{0.109}} & \textit{\textbf{0.163}} & \textit{\underline{0.254}} & \textit{\textbf{0.282}} & \textit{\textbf{0.324}} & \textit{\underline{0.359}}\\
		& MAE& \textit{\textbf{0.173}} & \textit{\textbf{0.206}} & \textit{\textbf{0.221}} & \textit{\underline{0.276}} & \textit{\textbf{0.194}} & \textit{\textbf{0.229}} & \textit{\textbf{0.257}} & \textit{\textbf{0.316}} & \textit{\textbf{0.363}} & \textit{\textbf{0.386}} & \textit{\textbf{0.417}} & \textit{\textbf{0.444}} \\
		\multirow{2}{*}{FEDformer}   & {MSE} & {0.103} & {0.129} & {0.132} & {0.134} & {0.072} & {\textit{\underline{0.102}}} & {0.130} & {0.178} & {\textbf{\textit{0.253}}} & {\textit{\underline{0.282}}} & {0.346} & {0.422} \\
		& {MAE} & {0.252} & {0.285} & {0.291} & {0.293} & {0.206} & {\textit{\underline{0.245}}} & {\textit{\underline{0.279}}} & {\textit{\underline{0.325}}} & {\textit{\underline{0.370}}} & {\textit{\underline{0.386}}} & {0.431} & {0.484}
		 \\
		\multirow{2}{*}{Pyraformer}  & MSE& 0.143 & 0.159 & 0.196 & 0.230 & 0.461 & 0.781 & 1.372 & 5.780 & 0.347 & 0.436 & 0.493 & 0.614 \\
		& MAE& 0.309 & 0.322 & 0.372 & 0.410 & 0.527 & 0.683 & 0.913 & 1.878 & 0.432 & 0.493 & 0.526 & 0.605 \\
		\multirow{2}{*}{ETSformer}   & MSE& \textit{\underline{0.060}} & \textit{\underline{0.081}} & {{0.098}} & {{0.119}} & 0.080 & 0.110 & 0.136 & 0.185 & 0.726 & 0.667 & 0.770 & 0.766 \\
		& MAE& \textit{\underline{0.190}} & \textit{\underline{0.221}}& {{0.248}} & {{0.282}} & 0.213 & 0.252 & 0.283 & 0.333 & 0.656 & 0.625 & 0.677 & 0.674 \\
		{}                            & {MSE} & {0.169} & {0.215} & {0.252} & {0.336} & {0.094} & {0.129} & {0.167} & {0.246} & {0.508} & {0.506} & {0.531} & {0.583} \\
		\multirow{-2}{*}{{AirFormer}} & {MAE} & {0.337} & {0.385} & {0.423} & {0.509} & {0.236} & {0.279} & {0.322} & {0.395} & {0.543} & {0.534} & {0.548} & {0.583}\\		
		\multirow{2}{*}{SCINet}      & MSE& 0.119 & 0.129 & 0.160 & 0.243 & 0.076 & 0.102 & \textit{\underline{0.129}} & \textit{\underline{0.176}} & 0.312 & 0.314 & \textit{\underline{0.332}} & 0.364 \\
		& MAE& 0.269 & 0.280 & 0.322 & 0.414 & 0.210 & 0.248& 0.280 & 0.328& 0.411 & 0.416 &\textit{\underline{0.427}} & 0.451 \\
		\multirow{2}{*}{TS2Vec}      & MSE& 0.098 & 0.153 & 0.169 & 0.164 & 0.088 & 0.122 & 0.158 & 0.200 & 0.315 & 0.333 & 0.347 & \textit{\textbf{0.350}} \\
		& MAE& 0.241 & 0.302 & 0.326 & 0.327 & 0.224 & 0.271 & 0.314 & 0.357 & 0.419 & 0.430 & 0.440 & \textit{\underline{0.447}} \\
		\multirow{2}{*}{DLinear}     & MSE& 0.111 & 0.136 & 0.166 & 0.280 & 0.094 & 0.130 & 0.164 & 0.223 & 0.411 & 0.385 & 0.410 & 0.447 \\
		& MAE& 0.258 & 0.286 & 0.325 & 0.453 & 0.237 & 0.278 & 0.316 & 0.369 & 0.473 & 0.455 & 0.470 & 0.502 \\
		\multirow{2}{*}{N-HiTS}      & MSE& 0.144 & 0.172 & 0.178 & 0.291 & \textit{\underline{0.071}} & 0.113 & 0.164 & 0.226 & 0.328 & 0.343 & 0.395 & 0.449 \\
		& MAE& 0.308 & 0.338 & 0.342 & 0.463 & \textit{\underline{0.195}} & 0.251 & 0.314 & 0.374 & 0.405 & 0.412 & 0.449 & 0.489 \\
		\multirow{2}{*}{FiLM} & MSE & 0.066 & 0.083 & \textit{\underline{0.097}} & \textbf{\textit{0.102}} & 0.152 & 0.175 & 0.197 & 0.238 & 0.967 & 0.958 & 0.992 & 1.038 \\
		& MAE & 0.199 & 0.225 & \textit{\underline{0.248}} & \textbf{\textit{0.252}} & 0.304 & 0.325 & 0.347 & 0.387 & 0.795 & 0.788 & 0.799 & 0.818\\
		\midrule
		\midrule
		\multirow{2}{*}{Methods}     & \multirow{2}{*}{Metrics} & \multicolumn{4}{c|}{WTH}   & \multicolumn{4}{c|}{Traffic}         & \multicolumn{4}{c}{Exchange}        \\
		\cmidrule(lr){3-6}
		\cmidrule(lr){7-10}
		\cmidrule(lr){11-14}
		&   & 96& 192& 336& 720& 96& 192& 336& 720& 96& 192& 336& 720\\
		\midrule[1pt]
		\multirow{2}{*}{GBT-Vanilla} & MSE& \textit{\underline{0.188}} & \textit{\textbf{0.221}} & \textit{\textbf{0.239}} & \textit{\textbf{0.218}} & \textit{\textbf{0.133}} & \textit{\textbf{0.140}} & \textit{\textbf{0.138}} & \textit{\textbf{0.174}} & \textit{\textbf{0.100}} & \textit{\textbf{0.186}} & \textit{\underline{0.408}} & \textit{\underline{0.925}} \\
		& MAE& \textit{\underline{0.318}} & \textit{\textbf{0.348}} & \textit{\textbf{0.372}} & \textit{\textbf{0.349}} & \textit{\textbf{0.222}} & \textit{\textbf{0.228}} & \textit{\textbf{0.234}} & \textit{\textbf{0.268}} & \textit{\textbf{0.249}} & \textit{\textbf{0.343}} & 0.522 & \textit{\underline{0.743}} \\
		\multirow{2}{*}{FEDformer} & {MSE} & {0.233} & {0.291} & {0.318} & {0.331} & {\textit{\underline{0.207}}} & {\textit{\underline{0.205}}} & {\textit{\underline{0.219}}} & {\textit{\underline{0.244}}} & {0.154} & {0.286} & {0.511} & {1.301} \\
		& {MAE} & {0.353} & {0.406} & {0.422} & {0.432} & {\textit{\underline{0.312}}} & {\textit{\underline{0.312}}} & {\textit{\underline{0.323}}} & {\textit{\underline{0.344}}} & {0.304} & {0.420} & {0.555} & {0.879}
		 \\
		\multirow{2}{*}{Pyraformer}  & MSE& 0.213 & 0.262 & 0.303 & 0.398 & 0.501 & 0.541 & 0.557 & 0.596 & 0.627 & 1.010 & 1.227 & 1.742 \\
		& MAE& 0.342 & 0.383 & 0.415 & 0.483 & 0.512 & 0.532 & 0.541 & 0.561 & 0.639 & 0.820 & 0.915 & 1.134 \\
		\multirow{2}{*}{ETSformer}   & MSE& 0.243 & 0.296 & 0.339 & 0.432 & 0.243 & 0.241 & 0.240 & 0.252 & \textit{\underline{0.100}}& 0.226 & 0.434 & 0.990 \\
		& MAE& 0.363 & 0.400 & 0.430 & 0.492 & 0.355 & 0.352 & 0.353 & 0.362 & \textit{\underline{0.252}}& \textit{\underline{0.353}} & \textit{\textbf{0.500}} & 0.821 \\
		{}                            & {MSE} & {0.226} & {0.272} & {0.308} & {0.396} & {0.540} & {0.504} & {0.532} & {0.554} & {0.246} & {0.508} & {0.915} & {0.977} \\
		\multirow{-2}{*}{{AirFormer}} & {MAE} & {0.358} & {0.394} & {0.425} & {0.488} & {0.537} & {0.516} & {0.531} & {0.544} & {0.400} & {0.577} & {0.777} & {0.786}\\
		\multirow{2}{*}{SCINet}      & MSE& 0.213 & 0.255 & 0.287 & 0.352 & 0.217 & 0.299 & 0.259 & 0.278 & 0.209 & 0.347 & 0.575 & 1.378 \\
		& MAE& 0.341 & 0.375 & 0.399 & 0.449 & 0.330 & 0.397 & 0.365 & 0.379 & 0.366 & 0.475 & 0.604 & 0.939 \\
		\multirow{2}{*}{TS2Vec}      & MSE& 0.199 & 0.240 & \textit{\underline{0.262}} & \textit{\underline{0.281}} & 0.357 & 0.359 & 0.368 & 0.380 & 0.184 & 0.373 & 0.666 & 2.941 \\
		& MAE& 0.323 & 0.361 & \textit{\underline{0.384}} & \textit{\underline{0.405}} & 0.431 & 0.433 & 0.440 & 0.447 & 0.315 & 0.452 & 0.612 & 1.313 \\
		\multirow{2}{*}{DLinear}     & MSE& 0.207 & 0.257 & 0.293 & 0.378 & 0.361 & 0.309 & 0.305 & 0.351 & 0.118 & \textit{\underline{0.222}} & \textit{\textbf{0.400}} & \textit{\textbf{0.837}} \\
		& MAE& 0.336 & 0.376 & 0.402 & 0.470 & 0.442 & 0.395 & 0.392 & 0.425 & 0.277 & 0.382 & \textit{\underline{0.506}} & \textit{\textbf{0.722}} \\
		\multirow{2}{*}{N-HiTS}      & MSE& \textit{\textbf{0.183}} & \textit{\underline{0.227}} & 0.265 & 0.359 & 0.284 & 0.264 & 0.269 & 0.298 & 0.241 & 0.870 & 1.809 & 2.144 \\
		& MAE& \textit{\textbf{0.307}} & \textit{\underline{0.352}} & 0.384 & 0.462 & 0.369 & 0.354 & 0.361 & 0.384 & 0.372 & 0.700 & 1.061 & 1.152  \\
		\multirow{2}{*}{FiLM} & MSE & 0.206 & 0.264 & 0.309 & 0.410 & 1.861 & 1.845 & 1.836 & 1.828 & 0.152 & 0.256 & 0.463 & 1.002 \\
		& MAE & 0.322 & 0.370 & 0.405 & 0.475 & 1.173 & 1.169 & 1.167 & 1.163 & 0.308 & 0.406 & 0.531 & 0.774\\
		\bottomrule[1.5pt]      
	\end{tabular}
\end{table}
\label{setting}
Details of hyper-parameters/settings of experiments are shown in Table \ref{taba3}. Most of hyper-parameters are set identically to those commonly used ones. The number of pyramid networks and the number of AR Blocks in the main pyramid network are all set to 3 according to \citet{RTNet} and \citet{informer} in that pyramid networks used in the first stage of GBT are motivated from them. In other experiments when GBT is compared with other methods or combined with other baselines, we reference corresponding hyper-parameters/settings to ensure that experiment results are obtained under the same circumstances for persuasive comparison.\par
$MSE={\sum{(x_i-\hat{x}_i)}^2}/{n}$ and $MAE={\sum{|x_i-\hat{x}_i|}}/{n}$ are used as metrics. All experiments are repeated for 5 times except special instruction and their average results are taken as final results. During multivariate forecasting of eight benchmark datasets, variates are treated as relevant variates only in WTH and are considered as different instances in other datasets {\citep{TS2Vec}}. Most of results of these baselines are directly taken from their papers if {existing experiment results with input sequence length of 96} and we supplement rest of experiments using their default settings {except input sequence length is fixed as 96. Note that we would rather redo experiments of many chosen baselines in that their results in the original papers are produced by the models with different input sequence lengths. However, \citet{Autoformer} points out that input sequence length counts tremendously for the prediction performances. Therefore, we fix the input sequence length of all baselines as 96, which is the commonly agreed-upon number proposed by \citet{Autoformer} (36 for ILI dataset following the same research).} `---' means that models fail for the out-of-memory (24GB) even when the size of batch is 1. {\textbf{\textit{The best results are highlighted in bold and italic}}} while {\textit{\underline{the second best results are highlighted in underline and italic}}} except specific instructions.\par
\subsection{Main Results}
\label{main results}
We compare forecasting capabilities of GBT with most of baselines and datasets aforementioned. We choose a commonly used prediction length group {\citep{Autoformer, ETSformer, FEDformer}}, i.e., \{96, 192, 336, 720\}, to perform multivariate/univariate forecasting experiments. In these experiments, as GBT only employs the canonical self-attention selected from Vanilla Transformer {\citep{attention2017}}, it is named as `GBT-Vanilla'. {The results of FEDformer and DLinear are separately the results of FEDformer-f and DLinear-S, which are their default and relatively better versions in general. The full results of their multiple versions, i.e., \{FEDformer-f, FEDformer-w\} and \{DLinear-S, DLinear-I\}, are presented in Appendix \ref{full_result}. The results of GBT-Vanilla in this section are also presented there for further comparison. Note that DLinear-I and DLinear-S are identical under univariate forecasting conditions \citep{DLinear}.}\par
We remove results of few datasets and baselines in this section to avoid excessive stacking of data. Concrete reasons are shown as below:
\begin{enumerate}
	\item We choose WTH in \{WTH, weather\} in that they are all datasets about climate and WTH is broader. However, we still provide forecasting results of all baselines on weather in Appendix \ref{weather}.
	\item We abandon ILI here in that it is a quite small dataset compared with others and has clear seasonal terms for all variates {\citep{N-HiTS}}, which means that non-stationarity is not apparent in ILI. Note that other datasets all contain non-stationary time series. Thus, GBT will not be very effective in improving forecasting capabilities of TSFTs on ILI so that performances of GBT-Vanilla is not satisfactory. However, we will show that GBT combined with frequence-enhanced methods, e.g., Autoformer {\citep{Autoformer}}, can still achieve state-of-the-art performances in Appendix \ref{ili}.
	\item Results of Informer, Autoformer and N-BEATS are removed in this section as their forecasting capabilities have been proved to be worse than at least half of the rest baselines. However, they are still very typical models, hence we will perform other types of experiments with them in later sections.
\end{enumerate}
\begin{table}[pos=b]
	\footnotesize
	\renewcommand{\arraystretch}{1.1}
	\centering
	\setlength\tabcolsep{4pt}
	\renewcommand{\multirowsetup}{\centering}
	\caption{Results of ablation study on GBT-Vanilla}
	\label{taba7}
	\begin{tabular}{c|c|cccc|cccc}
		\toprule[1.5pt]		
		\multirow{2}{*}{Methods}  & \multirow{2}{*}{Metrics} & \multicolumn{4}{c|}{ETTh$ _{1} $ (Univariate)}   & \multicolumn{4}{c}{ETTh$ _{1} $ (Multivariate)}  \\
		\cmidrule(lr){3-6}
		\cmidrule(lr){7-10}
		&    & 96      & 192   & 336   & 720   & 96        & 192   & 336   & 720   \\
		\midrule[1pt]
		\multirow{2}{*}{GBT-first}& MSE    & \textit{\underline{0.060}}    &\textit{\underline{0.077}} & 0.096 & \textit{\underline{0.128}} & 0.406     & 0.472 & 0.515 & 0.555 \\
		& MAE    & \textit{\underline{0.186}}   & 0.210  & 0.240  & \textit{\underline{0.282}} & 0.419     & 0.461 & 0.471 & \textit{\underline{0.512}} \\
		\midrule
		\multirow{2}{*}{GBT-second}         & MSE    & \textbf{\textit{0.051}}   & \textbf{\textit{0.074}}&\textit{\textbf{0.080}} & \textbf{\textit{0.119}} & 0.398     & \textit{\underline{0.448}} & 0.497 & \textbf{\textit{0.538}} \\
		& MAE    & \textbf{\textit{0.173}}   &\textit{\textbf{0.206}} &\textit{\textbf{0.221}}& \textbf{\textit{0.276}} & \textit{\underline{0.418}}   & \textit{\underline{0.442}} & 0.470 & \textbf{\textit{0.505}} \\
		\midrule			
		\multirow{2}{*}{GBT-wo-simul} & MSE &0.092 &0.167 &0.189 &0.284 &0.412 &0.449 &0.500 &\textit{\underline{0.543}} \\
		& MAE&0.237 &0.331 &0.342 &0.455 &0.431 &0.448 &0.476 &0.521 \\
		\midrule
		\multirow{2}{*}{GBT-wo-ESM}         & MSE    & 0.061   & 0.076 & \textit{\underline{0.083}} & 0.134 & 0.407     & 0.467 & 0.483 & 0.564 \\
		& MAE    & 0.186   & \textit{\underline{0.209}} & \textit{\underline{0.223}} & 0.304 & 0.420     & 0.458 & \textit{\underline{0.459}} & 0.512 \\
		\midrule
		\multirow{2}{*}{GBT-wo-CB}   & MSE    & 0.085   & 0.079 & 0.149 & 0.246 & \textbf{\textit{0.381}}     & \textbf{\textit{0.431}} & \textit{\textbf{0.474}}& 0.614 \\
		& MAE    & 0.223   & 0.213 & 0.317 & 0.420 & \textbf{\textit{0.401}}     & \textbf{\textit{0.422}} &\textit{\textbf{0.452}}& 0.523 \\
		\midrule
		\multirow{2}{*}{GBT-wo-Pyra} & MSE    & 0.064   & 0.078 & 0.098 & 0.156 & 0.416     & 0.478 & \textit{\underline{0.488}} & 0.563 \\
		& MAE    & 0.193   & 0.212 & 0.243 & 0.326 & 0.431     & 0.468 & 0.468 & 0.516\\
		\bottomrule[1.5pt]
	\end{tabular}
\end{table}
\begin{table}[pos=b]
	\footnotesize
	\renewcommand{\arraystretch}{1.1}
	\centering
	\setlength\tabcolsep{4pt}
	\renewcommand{\multirowsetup}{\centering}	
	\caption{Results of GBT with cross-attention or start token}
	\label{tab3}
	\begin{tabular}{c|c|cccc|cccc}
		\toprule[1.5pt]		
		\multirow{2}{*}{Methods}  & \multirow{2}{*}{Metrics} & \multicolumn{4}{c|}{ETTh$ _{1} $ (Univariate)}   & \multicolumn{4}{c}{ETTh$ _{1} $ (Multivariate)}  \\
		\cmidrule(lr){3-6}
		\cmidrule(lr){7-10}
		&    & 96      & 192   & 336   & 720   & 96        & 192   & 336   & 720   \\
		\midrule[1pt]
		\multirow{2}{*}{GBT-second}         & MSE    & \textbf{\textit{0.051}}   & 0.074&\textit{\underline{0.080}} & \textbf{\textit{0.119}} & \textit{\textbf{0.398}}     & \textit{\textbf{0.448}} & \textit{\underline{0.497}} & \textbf{\textit{0.538}} \\
		& MAE    & \textbf{\textit{0.173}}   &\textit{\underline{0.206}} &\textit{\underline{0.221}}& \textbf{\textit{0.276}} & \textit{\textbf{0.418}}   & \textit{\textbf{0.442}} & 0.470 & \textbf{\textit{0.505}} \\
		\midrule
		\multirow{2}{*}{GBT-w-cross}&MSE&	\textit{\underline{0.061}} &	\textbf{\textit{0.073}} &	0.095 &	0.235 &	0.403& 	\textit{\underline{0.449}} &	\textbf{\textit{0.465}} &	0.584 \\
		&MAE&	\textit{\underline{0.187}} &	\textbf{\textit{0.204}} &	0.238 &	0.408 &	0.420 &	\textit{\underline{0.442}} &	\textbf{\textit{0.442}} &	\textit{\underline{0.515}} \\
		\midrule		
		\multirow{2}{*}{GBT-w-st}&	MSE&	0.063&	\textit{\underline{0.074}}&	\textbf{\textit{0.076}}&	\textit{\underline{0.166}}&	0.404&	0.463&	0.498&	\textit{\underline{0.582}} 	\\		
		&MAE&	0.192&	0.207&	\textbf{\textit{0.221}}&	\textit{\underline{0.331}}&	0.420 &	0.460& 	\textit{\underline{0.465}}&	0.534	\\
		\bottomrule[1.5pt]
	\end{tabular}
\end{table}

It could be observed from Table \ref{tab1}/\ref{tab2} that GBT outperforms other baselines in most of univariate/multivariate forecasting situations even when it only employs the canonical attention mechanism. When compared with  FED-former/Pyraformer/ETSformer/AirFormer/SCINet/TS2Vec/DLinear/N-HiTS/FiLM, GBT-Vanilla yields {15.1\%}/52.6\%/ 10.6\%/61.1\%/32.3\%/ 42.1\%/{21.5\%}/12.6\%/46.5\% relative MSE reduction during multivariate forecasting and yields {23.4\%}/58.0\%/27.4\%/48.1\%/28.9\%/33.5\%/31.8\%/37.9\%/67.5\% relative MSE reduction during univariate forecasting in general.\par
\subsection{Ablation Study}
In this sub-section, ablation study is conducted to examine functions of two-stage framework, ESM, ConvBlock and pyramid architecture in GBT-Vanilla (GBT for short in the following). Six ablation variants of GBT are tested: 1) GBT-first: GBT only with the first stage; 2) GBT-second: GBT-Vanilla; 3) GBT-simul: GBT-Vanilla with simultaneously training of two stages. 4) GBT-wo-ESM: GBT without ESM; 5) GBT-wo-CB: GBT with feed-forward layers instead of ConvBlock in the first stage; 6) GBT-wo-Pyra: GBT without pyramid networks. Results of the second stage are shown except GBT-first in Table \ref{taba7}.\par
As Table \ref{taba7} shows, GBT-second performs better than other methods when forecasting under ETTh$_1$. GBT-Vanilla acquires 7.0\%/5.2\%/10.1\% relative MSE increase when we remove the second stage/ESM module/pyramid architecture, 17.9\% relative MSE increase when we replace ConvBlock with feed-forward layer and 27.6\% relative MSE increase when we simultaneously training both two stages of GBT. It means that without any of these four necessary components, GBT will suffer from worse performance due to diverse negative effects caused by them. This proves benefits of all components of GBT proposed by us. Specially, GBT-simul has the worse performance among all variants, illustrating that two-stage architecture is the core of GBT. \par
Apart from component contained in GBT, we extra examine whether cross-attention module and start token are redundant in \textit{Self-Regression} stage of GBT. To employ cross-attention module, we extra employ a network which is the same as the network of the first stage into the second stage to acquire key and value generated from input sequences as parts of inputs of cross-attention module during the second stage. To employ start token, we concatenate start token with outputs of the first stage and treat the whole of them as inputs of the second stage. Therefore, two additional variants of GBT are experimented: 1) `GBT-w-cross' refers to GBT with cross-attention module; 2) `GBT-w-st' denotes GBT with start token; in Table \ref{tab3}.\par
As Table \ref{tab3} shows, after we employ cross-attention module/start token into the second stage, forecasting results get worse in most of conditions. In general, GBT-second acquires 12.3\%/6.8\% relative MSE increase after we employ cross-attention module/start token. Therefore, cross-attention and start token are redundant and even do harm to performance of GBT in most of situations. \par
\subsection{Generality and Adaptivity Analysis}
\label{adaptivity}
\begin{table}
	\footnotesize
	\renewcommand{\arraystretch}{1.1}
	\centering
	\setlength\tabcolsep{4pt}
	\renewcommand{\multirowsetup}{\centering}		
	\caption{Results of GBT combined with other baselines}
	\label{taba8}
		\begin{tabular}{c|c|cccc|cccc}
			\toprule[1.5pt]
			\multirow{2}{*}{Methods}     & \multirow{2}{*}{Metrics} & \multicolumn{4}{c|}{WTH (Univariate)}& \multicolumn{4}{c}{WTH (Multivariate)}    \\
			\cmidrule(lr){3-6}
			\cmidrule(lr){7-10}
			&    & 96      & 192   & 336   & 720   & 96        & 192   & 336   & 720   \\
			\midrule[1.5pt]		
			\multirow{2}{*}{GBT+FEDformer} & MSE    & \textbf{\textit{0.171}}   & \textbf{\textit{0.200}}   & \textbf{\textit{0.204}} & \textbf{\textit{0.202}} & \textbf{\textit{0.449}}   & \textbf{\textit{0.489}} & \textbf{\textit{0.504}} & \textbf{\textit{0.508}} \\
			& MAE    & \textbf{\textit{0.303}}   & \textbf{\textit{0.331}} & \textbf{\textit{0.334}} & \textbf{\textit{0.332}} & \textbf{\textit{0.473}}   & \textbf{\textit{0.500}}   & \textbf{\textit{0.510}}  & \textbf{\textit{0.515}} \\
			\midrule
			\multirow{2}{*}{GBT+ETSformer}   & MSE    & \textbf{\textit{0.202}}  & \textit{\textbf{0.227}} & \textbf{\textit{0.234}}& \textbf{\textit{0.231}} & \textbf{\textit{0.444}}   & \textbf{\textit{0.494}} & \textbf{\textit{0.506}} & \textbf{\textit{0.532}} \\
			& MAE    & \textbf{\textit{0.329}}   & \textbf{\textit{0.359}} & \textbf{\textit{0.350}}  & \textbf{\textit{0.369}} & \textbf{\textit{0.474}}  & \textbf{\textit{0.515}}& \textbf{\textit{0.520}} & \textbf{\textit{0.544}} \\
			\midrule
			\multirow{2}{*}{GBT+SCINet}      & MSE    & \textbf{\textit{0.200}}     & \textbf{\textit{0.230}} & \textbf{\textit{0.232}} &\textbf{\textit{0.212}} & \textbf{\textit{0.436}}   & \textbf{\textit{0.484}} & \textbf{\textit{0.507}} & \textbf{\textit{0.509}} \\
			& MAE    & \textbf{\textit{0.325}}   & \textbf{\textit{0.351}} & \textbf{\textit{0.352}} & \textbf{\textit{0.339}} & \textbf{\textit{0.471}}  & \textbf{\textit{0.504}} & \textbf{\textit{0.522}} & \textbf{\textit{0.519}}\\
			\midrule
			\multirow{2}{*}{GBT+DLinear} & MSE & \textbf{\textit{0.196}} & \textit{\textbf{0.235}} & \textit{\textbf{0.237}} & \textbf{\textit{0.225}} & \textbf{\textit{0.497}}& \textbf{\textit{0.547}} &\textbf{\textit{0.577}} &\textbf{\textit{0.627}}  \\
			& MAE & \textit{\textbf{0.324}} & \textit{\textbf{0.358}}& \textit{\textbf{0.359}} & \textbf{\textit{0.352}} & \textit{\textbf{0.509}} & \textbf{\textit{0.543}} & \textbf{\textit{0.568}} & \textbf{\textit{0.595}}\\
			\bottomrule[1.5pt]
		\end{tabular}
\end{table}
\begin{table}[pos=b]
	\footnotesize
	\renewcommand{\arraystretch}{1.1}
	\centering
	\setlength\tabcolsep{2pt}
	\renewcommand{\multirowsetup}{\centering}
	\caption{Comparison experiment with RevIN}
	\label{tab4}
	\begin{tabular}{c|c|cccc|cccc}
		\toprule[1.5pt]
		\multirow{2}{*}{Methods}     & \multirow{2}{*}{Metrics} & \multicolumn{4}{c|}{ETTh$_1$ (Multivariate)}& \multicolumn{4}{c}{ETTm$_1$ (Multivariate)}    \\
		\cmidrule(lr){3-6}
		\cmidrule(lr){7-10}
		&    & 168      & 336   & 720   & 960   & 48      & 96   & 288   & 672   \\
		\midrule[1.5pt]	
		\multirow{2}{*}{Informer}       & MSE    & 1.138     & 1.278 & 1.357 & 1.470 & 0.499     & 0.605 & 0.906 & 0.943 \\
		& MAE    & 0.853     & 0.909 & 0.945 & 0.990 & 0.486     & 0.554 & 0.738 & 0.760 \\
		\midrule
		\multirow{2}{*}{Informer+RevIN} & MSE    & \textit{\underline{0.655}}     & \textit{\underline{1.058}} &\textit{\underline{0.926}} & \textit{\underline{0.902}} & \textit{\underline{0.390}}   & \textit{\underline{0.405}} &\textit{\underline{0.563}} & \textit{\underline{0.663}} \\
		& MAE    & \textit{\underline{0.561}}    & \textit{\underline{0.758}} & \textit{\underline{0.717}} & \textit{\underline{0.715}}& \textit{\underline{0.391}}     & \textit{\underline{0.411}} & \textit{\underline{0.502}} & \textit{\underline{0.550}}\\
		\midrule
		\multirow{2}{*}{Informer+GBT}   & MSE    & \textbf{\textit{0.443}}    & \textbf{\textit{0.511}} & \textbf{\textit{0.555}} & \textbf{\textit{0.596}} & \textbf{\textit{0.268}}     & \textbf{\textit{0.308}} & \textbf{\textit{0.363}} & \textbf{\textit{0.429}}\\
		& MAE    & \textbf{\textit{0.440}}     & \textbf{\textit{0.473}} & \textbf{\textit{0.512}} & \textbf{\textit{0.528}} & \textbf{\textit{0.337}}    & \textbf{\textit{0.366}} & \textbf{\textit{0.397}} & \textbf{\textit{0.442}}\\
		\bottomrule[1.5pt]
	\end{tabular}
\end{table}

We combine GBT with FEDformer and ETSformer separately and perform experiments on WTH to estimate the generality and adaptivity of GBT with TSFTs. We also combine GBT with {DLinear and} SCINet to further illustrate that its combination is not limited to TSFTs. Their combination formats are shown in Appendix \ref{combination}. {\textbf{\textit{The results which outperform corresponding ones shown in Table \ref{tab1}/\ref{tab2} are highlighted in bold and italic.}}}\par
It could be observed from Table \ref{taba8} that forecasting results of these three models are greatly improved under both univariate and multivariate situations of WTH after combined with GBT. Though they could not challenge TS2Vec, whose performances rank second under the forecasting of WTH in Table \ref{tab1}/\ref{tab2} (GBT-Vanilla ranks first), in their initial formulas, they surpass TS2Vec after combined with GBT in most of forecasting situations under WTH. FEDformer/ETSformer/SCINet{/DLinear} separately gain 28.6\%/25.4\%/15.3\%{/10.8\%} relative MSE reductions after combined with GBT in general. Meanwhile, GBT+FEDformer even outperforms GBT-Vanilla, demonstrating that GBT is general and adaptive to be combined with other SOTA modified TSFTs to achieve superior forecasting accuracy in the future.\par
\begin{table}
	\footnotesize
	\renewcommand{\arraystretch}{1.1}
	\centering
	\setlength\tabcolsep{2pt}
	\renewcommand{\multirowsetup}{\centering}
	\caption{Comparison experiment with Non-stationary Transformer}
	\label{tab5}
	\begin{tabular}{c|c|cccc|cccc}
		\toprule[1.5pt]
		\multirow{2}{*}{Methods}         & \multirow{2}{*}{Metrics} & \multicolumn{4}{c|}{Exchange (Multivariate)}         & \multicolumn{4}{c}{weather (Multivariate)}\\
		\cmidrule(lr){3-6}
		\cmidrule(lr){7-10}
		& & 96  & 192 & 336 & 720 & 96  & 192 & 336 & 720 \\
		\midrule[1pt]
		\multirow{2}{*}{Informer}        & MSE & 0.847         & 1.204         & 1.672         & 2.478         & 0.300         & 0.598         & 0.578         & 1.059         \\
		& MAE & 0.752         & 0.895         & 1.036         & 1.310         & 0.384         & 0.544         & 0.523         & 0.741         \\
		\midrule
		\multirow{2}{*}{Informer+Non-stationary Transformer} & MSE & \underline{\textit{0.129}}    & \underline{\textit{0.251}}    & \underline{\textit{0.373}}    & \underline{\textit{1.229}}    & \underline{\textit{0.186}}    & \underline{\textit{0.259}}    & \underline{\textit{0.295}}    & \underline{\textit{0.361}}    \\
		& MAE & \underline{\textit{0.258}}    & \underline{\textit{0.354}}    & \underline{\textit{0.434}}    & \underline{\textit{0.795}}    & \underline{\textit{0.235}}    & \underline{\textit{0.292}}    & \underline{\textit{0.317}}    & \underline{\textit{0.362}}    \\
		\midrule
		\multirow{2}{*}{Informer+GBT}    & MSE & \textit{\textbf{0.090}} & \textit{\textbf{0.188}} & \textit{\textbf{0.294}} & \textit{\textbf{0.776}} & \textit{\textbf{0.169}} & \textit{\textbf{0.209}} & \textit{\textbf{0.268}} & \textit{\textbf{0.331}} \\
		& MAE & \textit{\textbf{0.218}} & \textit{\textbf{0.320}} & \textit{\textbf{0.406}} & \textit{\textbf{0.661}} & \textit{\textbf{0.226}} & \textit{\textbf{0.265}} & \textit{\textbf{0.313}} & \textit{\textbf{0.358}}\\
		\bottomrule[1.5pt]
	\end{tabular}
\end{table}
\begin{table}
	\footnotesize
	\renewcommand{\arraystretch}{1.1}
	\centering
	\setlength\tabcolsep{2pt}
	\renewcommand{\multirowsetup}{\centering}
	\caption{The training/inference time and GPU memory consumption during univariate forecasting under ETTh$_1$ }
	\label{taba4}
	\begin{tabular}{c|c|cccc}		
		\toprule[1.5pt]
		\multirow{2}{*}{Methods}         & \multirow{2}{*}{Metric}       & \multicolumn{4}{c}{ETTh$ _{1} $ (Univariate) Batch\_size = 16}\\
		\cmidrule(lr){3-6}
		&& 96      & 192& 336 & 720 \\
		\midrule[1pt]
		\multirow{3}{*}{FEDformer}& GPU Memory/MB&	6758&	6844&	7216&	14025			 \\   
		& Train time per epoch/s & 217.350  & 215.727      & 280.154       & 598.435       \\
		& Inference Time/s       & 27.500    & 28.483       & 30.131        & 41.691        \\
		\midrule
		\multirow{3}{*}{GBT+FEDformer} & GPU Memory/MB&	\textbf{\textit{2922}}	&\textbf{\textit{3528}}	&\textbf{\textit{3992}}	&\textbf{\textit{5341}}
		\\ & Train time per epoch/s & \textbf{\textit{21.077+7.344}}      & \textbf{\textit{21.037+9.233}} & \textbf{\textit{20.981+12.885}} & \textbf{\textit{20.642+26.844}} \\
		& Inference Time/s       & \textbf{\textit{1.400+0.367}}       & \textbf{\textit{1.442+0.560}}  & \textbf{\textit{1.420+1.010}}   & \textbf{\textit{1.308+3.103}}   \\
		\midrule
		\multirow{3}{*}{Autoformer}& GPU Memory/MB&	3358	&3920	&4560	&5938\\
		& Train time per epoch/s & 46.326  & 50.998       & 56.738        & 76.022        \\
		& Inference Time/s       & 24.229  & 25.214       & 26.755        & 30.282        \\
		\midrule
		\multirow{3}{*}{GBT+Autoformer}& GPU Memory/MB	&\textbf{\textit{3186}}	&\textbf{\textit{3520}}	&\textbf{\textit{3992}}	&\textbf{\textit{5342}}\\
		& Train time per epoch/s & \textbf{\textit{13.271+7.599}}      & \textbf{\textit{13.496+9.287}} & \textbf{\textit{13.668+12.993}} & \textbf{\textit{13.922+27.340}} \\
		& Inference Time/s       & \textbf{\textit{1.939+0.277}}       & \textbf{\textit{1.926+0.583}}  & \textbf{\textit{1.794+1.010}}   & \textbf{\textit{1.691+3.031}}  \\
		\bottomrule[1.5pt]
	\end{tabular}
\end{table}

\subsection{Ability to Handle Non-stationary Time Series}
We conduct comparison experiments with RevIN {\citep{RevIN}} and Non-stationary Transformer {\citep{non-stationaryTransformer}}, which are other two advanced methods to help forecasting models deal with non-stationary time series. Multivariate forecasting conditions of \{ETTh$ _{1} $, ETTm$ _{1} $\} and \{Exchange, weather\} are separately chosen to perform comparison experiments with RevIN and Non-stationary Transformer. We choose different datasets for these two methods in that chosen datasets are their respective experiment circumstances in their original papers. Informer is selected as the baseline and experiment settings are chosen according to their original papers for fair comparison. Comparison results are respectively shown in Table \ref{tab4} and \ref{tab5}. As Table \ref{tab4} shows, Informer+GBT and Informer+RevIN both outperform Informer. The general MSE of Informer reduces 31.4\% after RevIN is employed. However, it reduces 57.0\% after we employ GBT, which improves more. Meanwhile, it could be observed from Table \ref{tab5} that Informer+GBT and Informer+Non-stationary Transformer both surpass Informer. Though the general MSE of Informer reduces 63.1\% after combined with Non-stationary Transformer, Informer with GBT obtains MSE reduction in a larger margin, i.e., 70.5\%. Therefore, GBT is a better solution for TSFTs when handling non-stationary time series.\par
\subsection{Complexity Analysis}
\label{Complexity}
We choose FEDformer and Autoformer, as baselines to examine the computation efficiency of GBT. The training time of each epoch and the whole inference time of the test dataset during univariate forecasting under ETTh$_1$ are shown in Table \ref{taba4}. The batch size is set to 16. The training/inference time and GPU memory consumption of models combined with GBT are reported by two stages respectively. {\textbf{\textit{The results which outperform the corresponding ones are highlighted in bold and italic.}}} Apparently, the training time/inference time/GPU memory consumption of FEDformer and Autoformer are decreased by 88.2\%/92.0\%/53.0\% and 52.3\%/88.7\%/9.5\% separately after combined with GBT. It means that GBT could tremendously reduce time and space complexity of FEDformer/Autoformer.\par

\subsection{Robustness Analysis}
\begin{table}
	\footnotesize
	\renewcommand{\arraystretch}{1.1}
	\centering
	\setlength\tabcolsep{2pt}
	\renewcommand{\multirowsetup}{\centering}
	\caption{Means/Stds of MSEs under ETTm$_2$ }
	\label{tab6}	
	\begin{tabular}{c|c|cccccccc}
		\toprule[
		1.5pt]
		\multirow{2}{*}{Methods}&\multirow{2}{*}{Metric}& \multicolumn{2}{c}{96}    & \multicolumn{2}{c}{192}   & \multicolumn{2}{c}{336}   &  \multicolumn{2}{c}{720}        \\
		\cmidrule(lr){3-4}
		\cmidrule(lr){5-6}
		\cmidrule(lr){7-8}
		\cmidrule(lr){9-10}
		&& Mean  & Std   & Mean  & Std   & Mean  & Std   & Mean  & Std   \\
		\midrule[1pt]
		\multirow{2}{*}{FEDformer}     & MSE    & 0.085 & 0.005 & 0.135 & 0.009 & 0.223 & 0.035 & 0.306 & 0.015 \\
		& MAE    & 0.224 & 0.006 & 0.285 & 0.009 & 0.366 & 0.028 & 0.439 & 0.012 \\
		\midrule
		\multirow{2}{*}{GBT+FEDformer} & MSE    & \textbf{\textit{0.069}} & \textbf{\textit{0.001}} & \textbf{\textit{0.105}}& \textbf{\textit{0.001}} & \textbf{\textit{0.139}} & \textbf{\textit{0.002}}& \textbf{\textit{0.177}} & \textbf{\textit{0.004}} \\
		& MAE    & \textbf{\textit{0.189}} & \textbf{\textit{0.001}} & \textbf{\textit{0.240}} & \textbf{\textit{0.002}} & \textbf{\textit{0.283}}& \textbf{\textit{0.002}} & \textbf{\textit{0.326}} & \textbf{\textit{0.008}} \\
		\midrule
		\multirow{2}{*}{Autoformer}      & MSE    & 0.213 & 0.059 & 0.348 & 0.227 & 0.435 & 0.337 & 0.44  & 0.119 \\
		& MAE    & 0.355 & 0.053 & 0.444 & 0.144 & 0.522 & 0.223 & 0.547 & 0.083 \\
		\midrule
		\multirow{2}{*}{GBT+Autoformer}  & MSE    & \textbf{\textit{0.080}} & \textbf{\textit{0.002}} & \textbf{\textit{0.113}} & \textbf{\textit{0.002}} & \textbf{\textit{0.138}} & \textbf{\textit{0.001}} & \textbf{\textit{0.185}} & \textbf{\textit{0.002}} \\
		& MAE    &\textbf{\textit{0.214}} & \textbf{\textit{0.003}} & \textbf{\textit{0.255}} & \textbf{\textit{0.002}} & \textbf{\textit{0.288}}& \textbf{\textit{0.003}}& \textbf{\textit{0.334}} & \textbf{\textit{0.002}}\\
		\bottomrule[1.5pt]
	\end{tabular}
\end{table}

Here we define that a model is more robust only when its forecasting result is less influenced by weight initialization. According to former analysis, TSFTs are easier to suffer from {the} over-fitting when handling non-stationary time series, i.e., they are less robust. So, we conduct this sub-experiment to check whether GBT could ameliorate this situation under univariate forecasting condition of ETTm$ _{2} $. The prediction length is chosen within \{96, 192, 336, 720\} and each experiment is conducted for 20 times while means and standard deviations (stds) of their prediction MSEs are shown in Table \ref{tab6}. {\textbf{\textit{The results which outperform the corresponding ones are highlighted in bold and italic.}}} As Table \ref{tab6} shows, MSEs of FEDformer/Autoformer own large stds/means. It means that these baselines are easy to be affected by random weight initialization. In contrast, stds/means of them drop sharply and are more steady after combined with GBT. It shows that GBT could enhance the robustness of TSFTs. Meanwhile, it mediately illustrates the seriousness of zero-initialization problem and necessity to solve it.\par

\section{Conclusion}
In this paper, we point out and analyze {the} over-fitting problem of time series forecasting Transformer caused by zero-initialization of decoder inputs {especially} when handling non-stationary time series. We propose GBT as a feasible and efficient solution. The two-stage framework of GBT decouples \textit{Auto-Regression} and \textit{Self-Regression} components of TSFTs to handle the problem brought by different statistical properties of input/prediction windows and zero-initialization of decoder inputs. We also propose {the} \textbf{\textit{E}}rror \textbf{\textit{S}}core \textbf{\textit{M}}odification module to deal with the `\textit{Good Beginning}' of the second stage in a more appropriate way. Extensive experiments on real-world datasets illustrate the promising forecasting capability of GBT. Furthermore, GBT is adaptive enough to easily couple with most of state-of-the-art forecasting models, {especially} TSFTs.\par
\section*{Acknowledgment}
This work was partially supported by National Natural Science Foundation of China under grant \#U19B2033, the National Key Research and Development Program of China under grant No. 2022YFB3904303 and the National Natural Science Foundation of China under grant No. 62076019. \par
\clearpage
\appendix
\numberwithin{figure}{section}
\numberwithin{table}{section}
\numberwithin{equation}{section}
\section{Zero-initialization Considering Position Embedding}
We analyze zero-initialization of prediction elements in decoders ignoring position embedding and bias in the main text to prove that start token could not help initialize prediction elements. Even if considering position embedding of TSFTs, this will not change this condition and we prove it as below.\par
Here we check the first masked self-attention module in decoders again {as Equation \ref{eq_attn1}}. $Q_i=Q_{i1}+Q_{i2}, K_i=K_{i1} +K_{i2} (i=1,2)$, where $Q_{i1}/K_{i1}$ denote the latent representations of \{start token, prediction elements\} in query/key while $Q_{i2}/K_{i2}$ are the latent representations of position embedding in them. Similarity, $Q_{21}/K_{21}$ will be zero tensors. Then we get query-key matching matrix as Equation \ref{eqa1}. Though zero tensors do not exist in query-key matching matrix, models obtain representations just through start token and position embedding.  Therefore, models could only learn position differences of prediction elements but could not acquire the information of statistical properties of prediction elements by using start token, i.e., start token has limited effects on inference process of prediction elements when handling non-stationary time series.
\begin{eqnarray}
	\label{eq_attn1}	 
	{Attn=(Q,K,V) = \frac{QK^\top}{\sqrt{d}}V,\ \ Q=[Q_1,Q_2]^\top,\ \ K=[K_1,K_2]^\top }
\end{eqnarray}

\begin{eqnarray}
	\label{eqa1}
	\begin{split}
		QK^\top &=
		\begin{bmatrix}
			Q_1 \\
			Q_2  
		\end{bmatrix}
		\times 
		\begin{bmatrix}
			K_1 \\
			K_2  
		\end{bmatrix}^\top
		=\begin{bmatrix}
			Q_{11}+Q_{12} \\
			Q_{21}+Q_{22}  
		\end{bmatrix}
		\times 
		\begin{bmatrix}
			K_{11}+K_{12} \\
			K_{21}+K_{22}  
		\end{bmatrix}^\top\\
		&=\begin{bmatrix}
			Q_1 \\
			Q_{22}  
		\end{bmatrix}
		\times 
		\begin{bmatrix}
			K_1 \\
			K_{22}  
		\end{bmatrix}^\top= \begin{bmatrix}
			Q_1K_1^\top & Q_1K_{22}^\top \\
			Q_{22}K_1^\top & Q_{22}K_{22}^\top 
		\end{bmatrix}
	\end{split}
\end{eqnarray}	
\section{Brief Introductions to Baselines}
\label{baselines}
\textbf{Informer} {\citep{informer}}: Time series forecasting Transformer (TSFT) with ProbSparse attention, distilling operation and one-forward decoder.\par
\textbf{Autoformer} {\citep{Autoformer}}: TSFT with Auto-Correlation attention and deep season-trend input sequence decomposition.\par
\textbf{FEDformer} {\citep{FEDformer}}: TSFT with frequency enhanced decomposition based on both Fourier and Wavelet form.\par
\textbf{Pyraformer} {\citep{Pyraformer}}: TSFT with pyramid architecture and low-complexity attention.\par
\textbf{ETSformer} {\citep{ETSformer}}: TSFT with Exponential Smoothing (ES) and season-level-growth decomposition.\par
{Airformer \citep{AirFormer}: TSFT with dartboard spatial attention, causal temporal attention and variational autoencoder (VAE).}\par
\textbf{TS2Vec} {\citep{TS2Vec}}: Self-supervised forecasting method with dilated CNN to extract universal feature maps.\par
\textbf{SCINet} {\citep{scinet}}: Temporal convolution network with binary tree architecture for efficient forecasting.\par
\textbf{N-BEATS} {\citep{NBEATS}}: Simple forecasting method based on Linear Projection and season-trend decomposition.\par
\textbf{N-HiTS} {\citep{N-HiTS}}: Simple forecasting method based on Linear Projection and temporal interpolation method.\par
\textbf{DLinear} {\citep{DLinear}}: Simple forecasting method based on Linear Projection and moving average method.\par
{FiLM \citep{FiLM}: Frequency enhanced forecasting model with linear projection layers and Legendre Polynomials projection for reconstruction.}\par
\section{Combination of GBT with TSFTs}
\begin{figure*}
	\centering	
	\includegraphics[width=1\columnwidth]{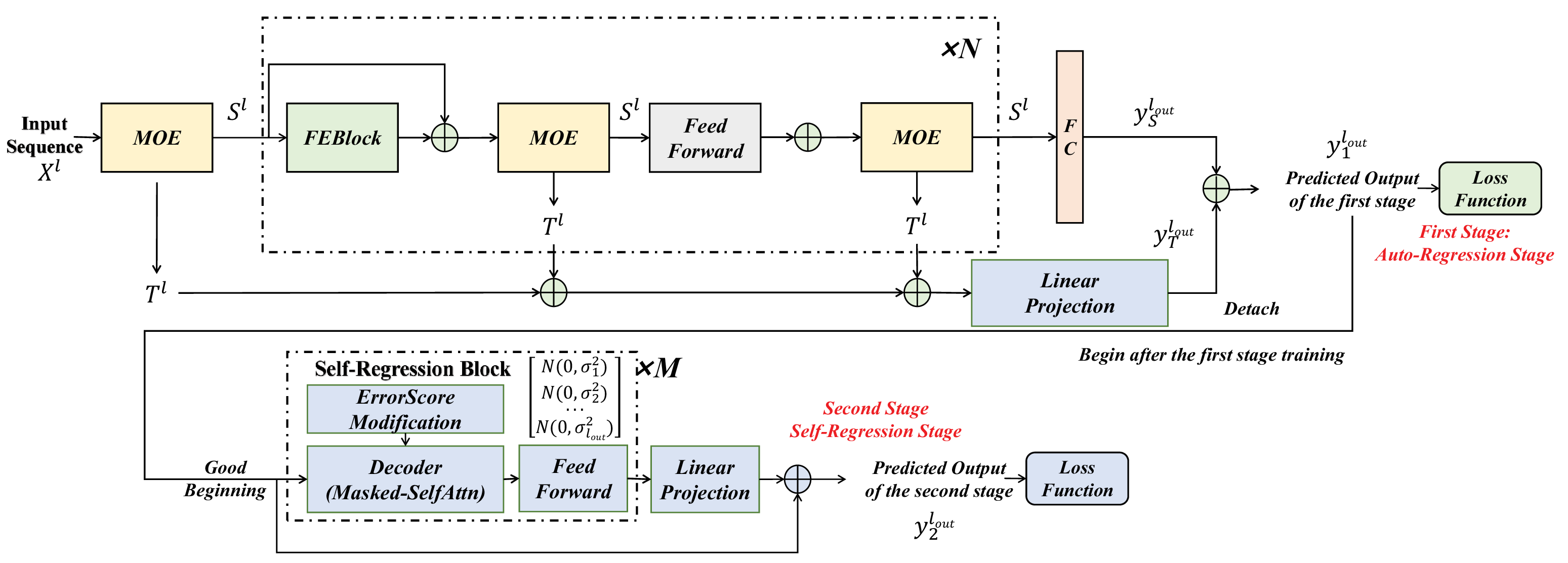}
	\caption{The architecture of FEDformer+GBT. The upper half is the \textit{Auto-Regression} stage including $N$ encoders of FEDformer and the lower half is the \textit{Self-Regression}		stage including $M$ \textit{Self-Regression} Blocks. MOE are used to decompose season and trend component from the input data and FEBlock are used to perform representation learning in frequency domain.}	
	\label{figa2}
\end{figure*}
\label{combination}
Here we introduce combinations of GBT with TSFTs and other SOTA time series forecasting models. As we mainly perform experiments on FEDformer, ETSformer and SCINet combined with GBT within Section \ref{adaptivity} in the main text, we present combinations of GBT with these three models. Combinations of GBT with other TSFTs or SOTA time series forecasting models are similar with them.\par
Following the seasonal-trend decomposition of FEDformer, the output of \textit{Auto-Regression} stage is also composed of season and trend components when GBT is combined with FEDformer as Figure \ref{figa2} shows. Trend and season components are both decomposed by MOE modules. In virtue of different latent dimensions, season/trend representations are projected through a FC/linear projection layer to obtain outputs of the \textit{Auto-Regression} stage. During \textit{Self-Regression} stage, the `\textit{Good Beginning}' is composed of season/trend components and we replace decoders of FEDformer with \textit{Self-Regression} Blocks used in GBT-Vanilla.\par
\begin{figure*}[pos=t]
	\centering	
	\includegraphics[width=0.8\columnwidth]{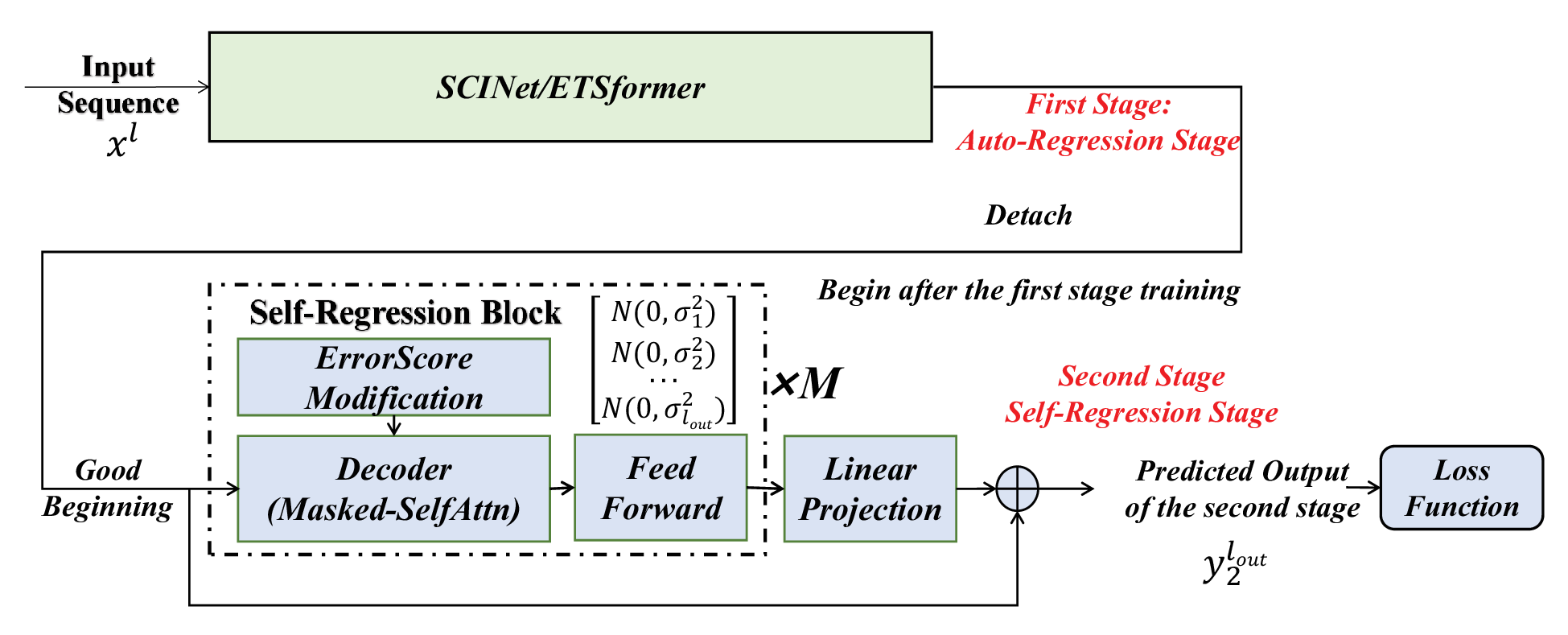}
	\caption{The architecture of ETSformer+GBT/SCINet+GBT. The upper half is the \textit{Auto-Regression} stage including networks of ETSformer/SCINet and the lower half is the \textit{Self-Regression} stage including $M$ \textit{Self-Regression} Blocks.}
	\label{figa3}
\end{figure*}

Combinations of GBT with ETSformer/SCINet are different in that these two baselines do not own any \textit{Self-Regression} module like masked self-attention mechanism. Therefore, whole networks of ETSformer/SCINet could be directly treated as networks in \textit{Auto-Regression} stage and we employ \textit{Self-Regression} Blocks used in GBT-Vanilla after their networks to combine them with GBT. Their architectures are shown in Figure \ref{figa3}.\par

If models {\citep{informer,Autoformer}} own \textit{Self-Regression} modules, e.g., masked self-attention mechanism, we choose the former format to combine them with GBT like combination of GBT with FEDformer. Otherwise, we choose another format to combine them with GBT like combinations of GBT with ETSformer/SCINet.\par
\section{Supplementary Experiment}
\subsection{Supplementary Experiment on Weather Dataset}
\label{weather}
Here we extra conduct multivariate/univariate forecasting experiments under weather dataset for more comprehensive comparison as Table \ref{tab7} shows. It is obvious that GBT-Vanilla outperforms other baselines in most of situations. N-HiTS surpasses GBT-Vanilla during univariate forecasting when input length is 96. However, when compared with FEDformer/Pyraformer/ETSformer/SCINet/TS2Vec/DLinear/N-HiTS, GBT-Vanilla yields 18.9\%/16.3\%/22.5\%/ 11.3\%/4.9\%/16.1\%/11.9\% relative MSE reduction during multivariate forecasting and 26.6\%/23.4\%/31.8\%/20.0\%/ 11.1\%/21.0\%/12.3\% relative MSE reduction during univariate forecasting in general.\par
\begin{table}
	\footnotesize
	\renewcommand{\arraystretch}{1.1}
	\centering
	\setlength\tabcolsep{2pt}
	\renewcommand{\multirowsetup}{\centering}
	\caption{Forecasting results under weather}
	\label{tab7}	
	\begin{tabular}{c|c|cccc|cccc}
		\toprule[1.5pt]
		\multirow{2}{*}{Methods}     & \multirow{2}{*}{Metrics} & \multicolumn{4}{c|}{weather (Multivariate)}    & \multicolumn{4}{c}{weather (Univariate)}      \\
		\cmidrule(lr){3-6}
		\cmidrule(lr){7-10}
		& & 96  & 192 & 336 & 720 & 96  & 192 & 336 & 720 \\
		\midrule[1pt]
		\multirow{2}{*}{GBT-Vanilla} & MSE & \textit{\textbf{0.434}} & \textit{\textbf{0.481}} & \textit{\textbf{0.514}} & \textit{\textbf{0.523}} & \underline{\textit{0.188}}    & \textit{\textbf{0.221}} & \textit{\textbf{0.239}} & \textit{\textbf{0.218}} \\
		& MAE & \textit{\textbf{0.466}} & \textit{\textbf{0.506}} & \textit{\textbf{0.527}} & \textit{\textbf{0.532}} & \underline{\textit{0.318}}    & \textit{\textbf{0.348}} & \textit{\textbf{0.372}} & \textit{\textbf{0.349}} \\
		\multirow{2}{*}{FEDformer}   & MSE & 0.531         & 0.601         & 0.646         & 0.631         & 0.236         & 0.289         & 0.332         & 0.335         \\
		& MAE & 0.525         & 0.564         & 0.618         & 0.597         & 0.358         & 0.407         & 0.431         & 0.437         \\
		\multirow{2}{*}{Pyraformer}  & MSE & 0.540         & 0.575         & 0.593         & 0.623         & 0.213         & 0.262         & 0.303         & 0.398         \\
		& MAE & 0.546         & 0.567         & 0.578         & 0.599         & 0.342         & 0.383         & 0.415         & 0.483         \\
		\multirow{2}{*}{ETSformer}   & MSE & 0.538         & 0.615         & 0.655         & 0.719         & 0.243         & 0.296         & 0.339         & 0.432         \\
		& MAE & 0.521         & 0.566         & 0.589         & 0.624         & 0.363         & 0.400         & 0.430         & 0.492         \\
		\multirow{2}{*}{SCINet}      & MSE & 0.489         & 0.526         & 0.572         & 0.617         & 0.213         & 0.255         & 0.287         & 0.352         \\
		& MAE & 0.495         & 0.524         & 0.562         & 0.586         & 0.341         & 0.375         & 0.399         & 0.449         \\
		\multirow{2}{*}{TS2Vec}      & MSE & \underline{\textit{0.450}}    & \underline{\textit{0.505}}    & \underline{\textit{0.532}}    & \underline{\textit{0.566}}    & 0.199         & 0.240         & 0.262         & \underline{\textit{0.281}}    \\
		& MAE & \underline{\textit{0.472}}    & \underline{\textit{0.515}}    & \underline{\textit{0.533}}    & \underline{\textit{0.557}}    & 0.323         & 0.361         & 0.384         & \underline{\textit{0.405}}    \\
		\multirow{2}{*}{DLinear}     & MSE & 0.514         & 0.572         & 0.597         & 0.646         & 0.207         & 0.257         & 0.293         & 0.378         \\
		& MAE & 0.514         & 0.553         & 0.570         & 0.603         & 0.336         & 0.376         & 0.402         & 0.470         \\
		\multirow{2}{*}{N-HiTS}      & MSE & 0.488         & 0.539         & 0.565         & 0.628         & \textit{\textbf{0.183}} & \underline{\textit{0.227}}    & \underline{\textit{0.265}}    & 0.359         \\
		& MAE & 0.496         & 0.536         & 0.555         & 0.596         & \textit{\textbf{0.307}} & \underline{\textit{0.352}}    & \underline{\textit{0.384}}    & 0.462  \\
		\bottomrule[1.5pt]
	\end{tabular}
\end{table}

\subsection{Supplementary Experiment on ILI Dataset}
\label{ili}
Though non-stationarity of time series in ILI is not distinct according to Section \ref{main results}, we additionally combine GBT with Autoformer/N-BEATS, which are separately frequence-enhanced TSFT/simple model, during multivariate/univariate forecasting under ILI dataset to show that GBT is still useful in relatively stationary forecasting conditions if combined with appropriate methods. Results are shown in Table \ref{tab8}. In this experiment, input length is set to 36 and prediction length group is \{24, 36, 48, 60\} according to \citet{Autoformer,ETSformer,FEDformer}. {\textbf{\textit{The results which outperform the corresponding original ones are highlighted in bold and italic}}}. It could be observed from Table \ref{tab8} that forecasting performances of N-BEATS and Autoformer are improved in all of situations. N-BEATS/Autoformer separately obtain 10.0\%/5.5\% relative MSE reduction during multivariate forecasting and 32.0\%/16.6\% relative MSE reduction during univariate forecasting in general.\par
\begin{table}
	\footnotesize
	\renewcommand{\arraystretch}{1.1}
	\centering
	\setlength\tabcolsep{2pt}
	\renewcommand{\multirowsetup}{\centering}
	\caption{Forecasting results under ILI}
	\label{tab8}	
	\begin{tabular}{c|c|cccc|cccc}
		\toprule[1.5pt]
		\multirow{2}{*}{Methods}        & \multirow{2}{*}{Metrics} & \multicolumn{4}{c|}{ILI (Multivariate)}    & \multicolumn{4}{c}{ILI (Univariate)}      \\
		\cmidrule(lr){3-6}
		\cmidrule(lr){7-10}
		& & 24  & 36  & 48  & 60  & 24  & 36  & 48  & 60  \\
		\midrule[1pt ]
		\multirow{2}{*}{N-BEATS}        & MSE & 5.294         & 5.435         & 6.145         & 5.976         & 2.010         & 2.705         & 2.059         & 2.657         \\
		& MAE & 1.582         & 1.684         & 1.797         & 1.780         & 1.039         & 1.259         & 1.130         & 1.243         \\
		\multirow{2}{*}{GBT+N-BEATS}    & MSE & \textit{\textbf{5.060}} & \textit{\textbf{4.807}} & \textit{\textbf{5.614}} & \textit{\textbf{5.050}} & \textit{\textbf{1.515}} & \textit{\textbf{1.820}} & \textit{\textbf{1.531}} & \textit{\textbf{1.464}} \\
		& MAE & \textit{\textbf{1.523}} & \textit{\textbf{1.539}} & \textit{\textbf{1.681}} & \textit{\textbf{1.576}} & \textit{\textbf{0.908}} & \textit{\textbf{1.007}} & \textit{\textbf{0.954}} & \textit{\textbf{0.946}} \\
		\multirow{2}{*}{Autoformer}     & MSE & 4.971         & 3.971         & 4.060         & 3.951         & 1.469         & 0.953         & 1.085         & 1.320         \\
		& MAE & 1.596         & 1.414         & 1.382         & 1.399         & 1.015         & 0.819         & 0.885         & 0.966         \\
		\multirow{2}{*}{GBT+Autoformer} & MSE & \textit{\textbf{4.351}} & \textit{\textbf{3.857}} & \textit{\textbf{3.935}} & \textit{\textbf{3.817}} & \textit{\textbf{1.087}} & \textit{\textbf{0.845}} & \textit{\textbf{0.918}} & \textit{\textbf{1.141}} \\
		& MAE & \textit{\textbf{1.465}} & \textit{\textbf{1.362}} & \textit{\textbf{1.334}} & \textit{\textbf{1.336}} & \textit{\textbf{0.844}} & \textit{\textbf{0.776}} & \textit{\textbf{0.811}} & \textit{\textbf{0.897}}\\
		\bottomrule[1.5pt]
	\end{tabular}
\end{table}
{\subsection{Comprehensive Comparison With DLinear Under Various Settings}
We conduct a comprehensive comparison of GBT-Vanilla and {DLinear-I, DLinear-S} with four ETT datasets under multivariate forecasting conditions. DLinear {\citep{DLinear}} empirically shows that its performance can be better than time series forecasting Transformers with different settings from the commonly agreed-upon one chosen by \citet{Autoformer, FEDformer, ETSformer}. This experiment is intended for showing that GBT still excels in handling these forecasting conditions, unlike other TSFTs. They are experimented with different input sequence lengths (in \{96, 336, 720\}) and different learning rates (in \{0.001, 0.0005, 0.0001\}). Ultimately, there are $ 3^2 = 9 $ settings for each forecasting condition. Only MSE is used as the evaluation metrics, otherwise the data would be exceedingly tremendous. The results are shown in Table \ref{tab_comparison}. The average and best results of the forecasting conditions with different prediction lengths and datasets for different baselines are also presented.\par
Although DLinear own two versions, either the best or the average performances of GBT-Vanilla are unequivocally better than those of DLinear in  the major forecasting conditions (Best performances: 14/16; Average performances: 15/16), demonstrating that the GBT architecture is more accurate and general. These phenomena justifiably verify that Transformer with proper usage literally pertains to time series forecasting, which is opposed to the opinion of DLinear.\par}

\begin{table}[]
	\scriptsize
	\renewcommand{\arraystretch}{1.1}
	\centering
	\setlength\tabcolsep{1.7pt}
	\renewcommand{\multirowsetup}{\centering}
	\caption{{Forecasting results of comprehensive comparison. All numbers are MSE results.}}
	\label{tab_comparison}
	\begin{tabular}{c|c|c|cccc|cccc|cccc|cccc}
		\toprule[1.5pt]
		
		\multirow{2}{*}{\makecell{Input\\Length}} & \multirow{2}{*}{\makecell{Learning\\Rate}} & \multirow{2}{*}{Methods} & \multicolumn{4}{c|}{ETTh$ _{1} $}       & \multicolumn{4}{c|}{ETTh$ _{2} $}       & \multicolumn{4}{c|}{ETTm$ _{1} $}       & \multicolumn{4}{c}{ETTm$ _{2} $}       \\
		\cmidrule(lr){4-7}
		\cmidrule(lr){8-11}
		\cmidrule(lr){12-15}
		\cmidrule(lr){16-19}
		&  &  & 96  & 192 & 336 & 720 & 96  & 192 & 336 & 720 & 96  & 192 & 336 & 720 & 96  & 192 & 336 & 720 \\
		\midrule[1pt]
		\multirow{9}{*}{96} & \multirow{3}{*}{1e-4}      & GBT-Vanilla    & \textit{\textbf{0.398}} & \textit{\textbf{0.448}} & \textit{\textbf{0.497}} & \textit{\textbf{0.538}} & \textit{\textbf{0.328}} & \textit{\textbf{0.468}} & \textit{\textbf{0.561}} & \textit{\textbf{0.691}} & \textit{\textbf{0.329}} & \textit{\textbf{0.366}} & \textit{\underline{0.413}}         & \textit{\textbf{0.454}} & \textit{\textbf{0.189}} & \textit{\textbf{0.249}} & \textit{\textbf{0.324}} & \textit{\textbf{0.395}} \\
		&  & DLinear-I      & 0.441         & 0.493         & 0.534         & \textit{\underline{0.553}}         & 0.651         & 0.894         & 1.081         & 1.291         & \textit{\underline{0.337}}        & \textit{\underline{0.373}}        & \textit{\textbf{0.405}} & \textit{\underline{0.467}}     & 0.254         & 0.432         & 0.648         & 0.986         \\
		&  & DLinear-S      & \textit{\underline{0.431}} & \underline{\textit{0.474}} & \textit{\underline{0.518}} & 0.560        & \textit{\underline{0.381}}         & \textit{\underline{0.508}}         & \textit{\underline{0.621}}        & \textit{\underline{0.856}}        & 0.350         & 0.387         & 0.417         & 0.477         & \textit{\underline{0.199}}         & \textit{\underline{0.287}}       & \textit{\underline{0.387}}        & \textit{\underline{0.544}}         \\
		\cmidrule(l){2-19}
		& \multirow{3}{*}{5e-4}      & GBT-Vanilla    & \textit{\textbf{0.380}} & \textit{\textbf{0.422}} & \textit{\textbf{0.461}} & \textit{\textbf{0.495}} & \textit{\textbf{0.304}} & \textit{\textbf{0.389}} & \textit{\textbf{0.446}} & \textit{\textbf{0.497}} &\textbf{\textit{0.327}} & \textit{\textbf{0.356}} & \textit{\underline{0.405}} & \textbf{\textit{0.446}} & \textit{\textbf{0.183}} & \textit{\textbf{0.266}} & \textit{\textbf{0.340}} & \textit{\textbf{0.450}} \\
		&  & DLinear-I      & 0.395         & 0.450         & 0.496         & \textit{\underline{0.511}}        & 0.567         & 0.801         & 1.010         & 1.196         & \textit{\underline{0.332}}         & \textit{\underline{0.370}}       & \textit{\textbf{0.401}}         & \textit{\underline{0.461}}        & 0.272         & 0.388         & 0.674         & 0.935         \\
		&  & DLinear-S      & \textit{\underline{0.392}}        & \textit{\underline{0.443}}       & \textit{\underline{0.491}}         & 0.519         & \textit{\underline{0.347}}      & \textit{\underline{0.470}}         & \textit{\underline{0.591}}         & \textit{\underline{0.791}}         & 0.346         & 0.383         & 0.417         & 0.475         & \textit{\underline{0.194}}         & \textit{\underline{0.284}}        & \textit{\underline{0.376}}        & \textit{\underline{0.506}}      \\
		\cmidrule(l){2-19}
		& \multirow{3}{*}{1e-3}      & GBT-Vanilla    & \textit{\textbf{0.390}} & \textit{\textbf{0.429}} & \textit{\textbf{0.468}} & 0.524         & \textit{\textbf{0.322}} & \textit{\textbf{0.410}} & \textit{\textbf{0.436}} & \textit{\textbf{0.535}} &\textit{\underline{0.337}} & \textit{\textbf{0.370}} & \textit{\underline{0.407}} & \textit{\textbf{0.447}}& \textit{\textbf{0.185}} & \textit{\underline{0.271}}        & \textit{\textbf{0.356}} & \textit{\textbf{0.491}} \\
		&  & DLinear-I      & 0.396         & 0.453         & 0.497         & \textit{\textbf{0.514}} & 0.564         & 0.780         & 0.958         & 1.295         & \textit{\textbf{0.334}}         & \textit{\underline{0.372}}      & \textit{\textbf{0.399}}        & \textit{\underline{0.467}  }       & 0.249         & 0.406         & 0.543         & 0.957         \\
		&  & DLinear-S      & \textit{\underline{0.392}}         & \textit{\underline{0.444}}        & \textit{\underline{0.485}}        & \textit{\underline{0.520}}         & \textit{\underline{0.332}}         & \textit{\underline{0.454}}        & \textit{\underline{0.570}}         & \textit{\underline{0.800}}         & 0.346         & 0.383         & 0.416         & 0.477         & \textit{\underline{0.188}}         & \textit{\textbf{0.264}} & \textit{\underline{0.381}}         & \textit{\underline{0.524}}       \\
		\midrule
		\multirow{9}{*}{336}& \multirow{3}{*}{1e-4}      & GBT-Vanilla    & \textit{\textbf{0.395}} & \textit{\underline{0.443}}         & \textit{\underline{0.476}}        & \textit{\textbf{0.502}} & \textit{\textbf{0.300}} & \textit{\textbf{0.360}} & \textit{\textbf{0.406}} & \textit{\textbf{0.532}} &\textit{\textbf{0.288}} & \textit{\textbf{0.323}} & \textit{\textbf{0.367}} & \textit{\textbf{0.415}}& \textit{\textbf{0.175}} & \textit{\textbf{0.229}} & \textit{\textbf{0.286}} & \textit{\textbf{0.403}} \\
		&  & DLinear-I      & 0.419         & 0.454         & 0.482         & 0.512         & 0.459         & 0.618         & 0.749         & 1.070         & \textit{\underline{0.293}}         & \textit{\underline{0.332}}        & \textit{\underline{0.371}}       & 0.438         & 0.198         & 0.316         & 0.528         & 0.803         \\
		&  & DLinear-S      & \textit{\underline{0.398}}         & \textit{\textbf{0.439}} & \textit{\textbf{0.463}} & \textit{\underline{0.504}}         & \textit{\underline{0.326}}         & \textit{\underline{0.414}}      & \textit{\underline{0.509}}       & \textit{\underline{0.771}}        & 0.303         & 0.339         & 0.377         &\textit{\underline{0.432}}       & \textit{\underline{0.175}}        & \textit{\underline{0.236}}         & \textit{\underline{0.321}}         & \textit{\underline{0.448}}         \\
		\cmidrule(l){2-19}
		& \multirow{3}{*}{5e-4}      & GBT-Vanilla    & \textit{\textbf{0.396}} & \textit{\underline{0.436}}         & \textit{\underline{0.460}}         & \textit{\textbf{0.490}} & \textit{\textbf{0.299}} & \textit{\textbf{0.366}} & \textit{\textbf{0.408}} & \textit{\textbf{0.568}} &\textit{\textbf{0.292}} & \textit{\textbf{0.325}} & \textit{\textbf{0.368}} & \textit{\textbf{0.416}}& \textit{\textbf{0.170}} & \textit{\textbf{0.223}} & \textit{\textbf{0.280}} & \textit{\textbf{0.399}} \\
		&  & DLinear-I      & \textit{\underline{0.388}}         & 0.436         & 0.461         & {{0.503}}         & 0.430         & 0.574         & 0.697         & 0.995         & \textit{\underline{0.292}}         & \textit{\underline{0.333}}         & \underline{\textit{0.373}}         & 0.432         & 0.215         & 0.299         & 0.547         & 0.790         \\
		&  & DLinear-S      & 0.388         & \textit{\textbf{0.423}} & \textit{\textbf{0.451}} & \textit{\underline{0.495}}         & \textit{\underline{0.301}}         & \textit{\underline{0.399}}         & \textit{\underline{0.469}}         & \textit{\underline{0.758}}         & 0.305         & 0.338         & 0.375         &\textit{\underline{0.431}}         & \textit{\underline{0.172}}         & \textit{\underline{0.244}}         & \textit{\underline{0.298}}         & \textit{\underline{0.420}}         \\
		\cmidrule(l){2-19}
		& \multirow{3}{*}{1e-3}      & GBT-Vanilla    & 0.402         & 0.434         & 0.474         & 0.534         & \textit{\underline{0.316}}         & \textit{\underline{0.405}}         & \textit{\textbf{0.444}} & \textit{\textbf{0.563}} &\textit{\textbf{0.299}} & \textit{\textbf{0.337}} & \textit{\textbf{0.373}} & \textit{\textbf{0.426}}& \textit{\underline{0.179}}        & \textit{\textbf{0.231}} & \textit{\textbf{0.287}} & \textit{\underline{0.420}}         \\
		&  & DLinear-I      & \textit{\textbf{0.382}} & \textit{\underline{0.433}}         & \textit{\underline{0.462}}         & \textit{\underline{0.516}}         & 0.403         & 0.537         & 0.677         & 1.015         & \textit{\underline{0.302}}         & 0.344         &\textit{\underline{0.373}}         & \textit{\underline{0.436}}         & 0.216         & 0.319         & 0.440         & 0.739         \\
		&  & DLinear-S      & \textit{\underline{0.387}}         & \textit{\textbf{0.425}} & \textit{\textbf{0.457}} & \textit{\textbf{0.492}} & \textit{\textbf{0.305}} & \textit{\textbf{0.389}} & \textit{\underline{0.458}}         & \textit{\underline{0.669}}         & 0.309         &\textit{\underline{0.337}}         & 0.377         & 0.441         & \textit{\textbf{0.174}} & \textit{\underline{0.242}}         & \textit{\underline{0.295}}         & \textit{\textbf{0.409}} \\
		\midrule
		\multirow{9}{*}{720}& \multirow{3}{*}{1e-4}      & GBT-Vanilla    & \textit{\underline{0.399}}         & \textit{\textbf{0.431}} & \textit{\underline{0.474}}         & \textit{\underline{0.535}}         & \textit{\textbf{0.293}} & \textit{\textbf{0.359}} & \textit{\textbf{0.426}} & \textit{\textbf{0.459}} &\textit{\underline{0.301}} & \textit{\textbf{0.320}} & \textit{\textbf{0.358}} & \textit{\underline{0.423}} & \textit{\underline{0.191}}         & \textit{\underline{0.244}}         & \textit{\underline{0.293}}         & \textit{\textbf{0.392}} \\
		&  & DLinear-I      & 0.417         & 0.453         & 0.478         & 0.537         & 0.455         & 0.657         & 0.794         & 1.293         & \textbf{\textit{0.300}}         & \textit{\underline{0.341}}         & 0.381         & 0.432         & 0.201         & 0.274         & 0.495         & 0.651         \\
		&  & DLinear-S      & \textit{\textbf{0.391}} & \textit{\underline{0.431}}         & \textit{\textbf{0.461}} & \textit{\textbf{0.506}} & \textit{\underline{0.332}}         & \textit{\underline{0.442}}         & \textit{\underline{0.547}}         & \textit{\underline{0.938}}         & 0.312         & 0.341         & \textit{\underline{0.375}}         & \textit{\textbf{0.422}}         & \textit{\textbf{0.175}} & \textit{\textbf{0.233}} & \textit{\textbf{0.292}} & \textit{\underline{0.418}}         \\
		\cmidrule(l){2-19}
		& \multirow{3}{*}{5e-4}      & GBT-Vanilla    & \textit{\underline{0.401}}         & \textit{\underline{0.447}}         & \textit{\underline{0.475}}         & 0.539         &\textit{\underline{0.340}}         & \textit{\textbf{0.404}} & \textit{\textbf{0.411}} & \textit{\textbf{0.498}} &\textit{\textbf{0.302}} & \textit{\textbf{0.330}} & \textit{\textbf{0.359}} & \textit{\textbf{0.424}}& \textit{\underline{0.187}}         & \textit{\underline{0.241}}         & \textit{\underline{0.308}}         & \textit{\underline{0.433}}         \\
		&  & DLinear-I      & 0.414         & 0.458         & 0.485         & \textit{\underline{0.531}}         & 0.420         & 0.622         & 0.725         & 1.334         & \textit{\underline{0.307}}         & \textit{\underline{0.342}}         & 0.387         & 0.445         & 0.199         & 0.275         & 0.369         & 0.579         \\
		&  & DLinear-S      & \textit{\textbf{0.392}} & \textit{\textbf{0.426}} & \textit{\textbf{0.461}} & \textit{\textbf{0.510}} & \textit{\textbf{0.309}} & \textit{\underline{0.428}}         & \textit{\underline{0.502}}         & \textit{\underline{0.824}}         & 0.315         & 0.345         & \textit{\underline{0.373}}         & \textit{\underline{0.429}}         & \textit{\textbf{0.167}} & \textit{\textbf{0.227}} & \textit{\textbf{0.292}} & \textit{\textbf{0.410}} \\
		\cmidrule(l){2-19}
		& \multirow{3}{*}{1e-3}      & GBT-Vanilla    & \textit{\underline{0.404}}         & \textit{\underline{0.450}}         & \textit{\textbf{0.475}} & \textit{\textbf{0.538}} & \textit{\underline{0.325}}         & \textit{\underline{0.417}}         & \textit{\textbf{0.425}} & \textit{\textbf{0.558}} &\textit{\textbf{0.278}} & \textit{\textbf{0.313}} & \textit{\textbf{0.356}} & \textit{\textbf{0.417}}& \textit{\textbf{0.171}} & \textit{\textbf{0.219}} & \textit{\textbf{0.272}} & \textit{\textbf{0.457}} \\
		&  & DLinear-I      & 0.406         & \textit{\textbf{0.448}} & 0.511         & 0.547         & 0.421         & 0.549         & 0.890         & 1.615         & 0.324         & \textit{\underline{0.353}}         & 0.394         & 0.469         & 0.232         & 0.319         & 0.446         & 0.641         \\
		&  & DLinear-S      & \textit{\textbf{0.393}} & 0.454         & \textit{\underline{0.475}}         & \textit{\underline{0.519}}         & \textit{\textbf{0.301}} & \textit{\textbf{0.371}} & \textit{\underline{0.561}}         & \textit{\underline{0.792}}         & \textit{\underline{0.322}}         & 0.354         & \textit{\underline{0.380}}         & \textit{\underline{0.427}}         & \textit{\underline{0.182}}         & \textit{\underline{0.238}}        & \textit{\underline{0.299}}         & \textit{\underline{0.464}}        \\
		\midrule
		\multicolumn{2}{c|}{\multirow{3}{*}{Best Results}} & GBT-Vanilla & \textit{\textbf{0.380}} & \textit{\textbf{0.422}} & \textit{\underline{0.460}}          & \textit{\textbf{0.490}} & \textit{\textbf{0.293}} & \textit{\textbf{0.359}} & \textit{\textbf{0.406}} & \textit{\textbf{0.459}} & \textit{\textbf{0.278}} & \textit{\textbf{0.313}} & \textit{\textbf{0.356}} & \textit{\textbf{0.415}} & \textit{\underline{0.170}}          & \textit{\textbf{0.219}} & \textit{\textbf{0.272}} & \textit{\textbf{0.392}} \\
		\multicolumn{2}{c|}{}                              & DLinear-I   & \textit{\underline{0.382}}          & 0.433                   & 0.461                   & 0.503                   & 0.403                   & 0.537                   & 0.677                   & 0.995                   & \textit{\underline{0.292}}          & \textit{\underline{0.332}}          & \textit{\underline{0.371}}          & 0.432                   & 0.198                   & 0.274                   & 0.369                   & 0.579                   \\
		\multicolumn{2}{c|}{}                              & DLinear-S   & 0.387                   & \textit{\underline{0.423}}          & \textit{\textbf{0.451}} & \textit{\underline{0.492}}          & \textit{\underline{0.301}}          & \textit{\underline{0.371}}          & \textit{\underline{0.458}}          & \textit{\underline{0.669}}          & 0.303                   & 0.337                   & 0.373                   & \textit{\underline{0.422}}          & \textit{\textbf{0.167}} & \textit{\underline{0.227}}          & \textit{\underline{0.292}}          & \textit{\underline{0.409}}         
		\\
		\midrule
		\multicolumn{2}{c|}{\multirow{3}{*}{Average Results}} & GBT-Vanilla & \textit{\textbf{0.396}} & \textit{\textbf{0.438}} & \textit{\textbf{0.473}} & \textit{\underline{0.522}} & \textit{\textbf{0.314}} & \textit{\textbf{0.398}} & \textit{\textbf{0.440}} & \textit{\textbf{0.544}} & \textit{\textbf{0.306}} & \textit{\textbf{0.338}} & \textit{\textbf{0.378}} & \textit{\textbf{0.430}} & \textit{\textbf{0.181}} & \textit{\textbf{0.241}} & \textit{\textbf{0.305}} & \textit{\textbf{0.427}} \\
		\multicolumn{2}{c|}{}                                 & DLinear-I   & 0.406         & 0.453                   & 0.490                   & 0.525                   & 0.485                   & 0.670                   & 0.842                   & 1.234                   & \textit{\underline{0.313}}          & \textit{\underline{0.351}}          & \textit{\underline{0.387}}          & 0.450                   & 0.226                   & 0.336                   & 0.521                   & 0.787                   \\
		\multicolumn{2}{c|}{}                                 & DLinear-S   & \textit{\underline{0.396}}          & \textit{\underline{0.440}}          & \textit{\underline{0.474}} & \textit{\textbf{0.514}}          & \textit{\underline{0.326}}          & \textit{\underline{0.431}}          & \textit{\underline{0.536}}          & \textit{\underline{0.800}}          & 0.323                   & 0.356                   & 0.390                   & \textit{\underline{0.446}}          & \textit{\underline{0.181}} & \textit{\underline{0.251}}          & \textit{\underline{0.327}}          & \textit{\underline{0.460}}         
		 
		\\
		\bottomrule[1.5pt]
	\end{tabular}
\end{table}
\subsection{Visualization of Error Accumulation}
\begin{figure}
	\centering	
	\includegraphics[width=0.6\columnwidth]{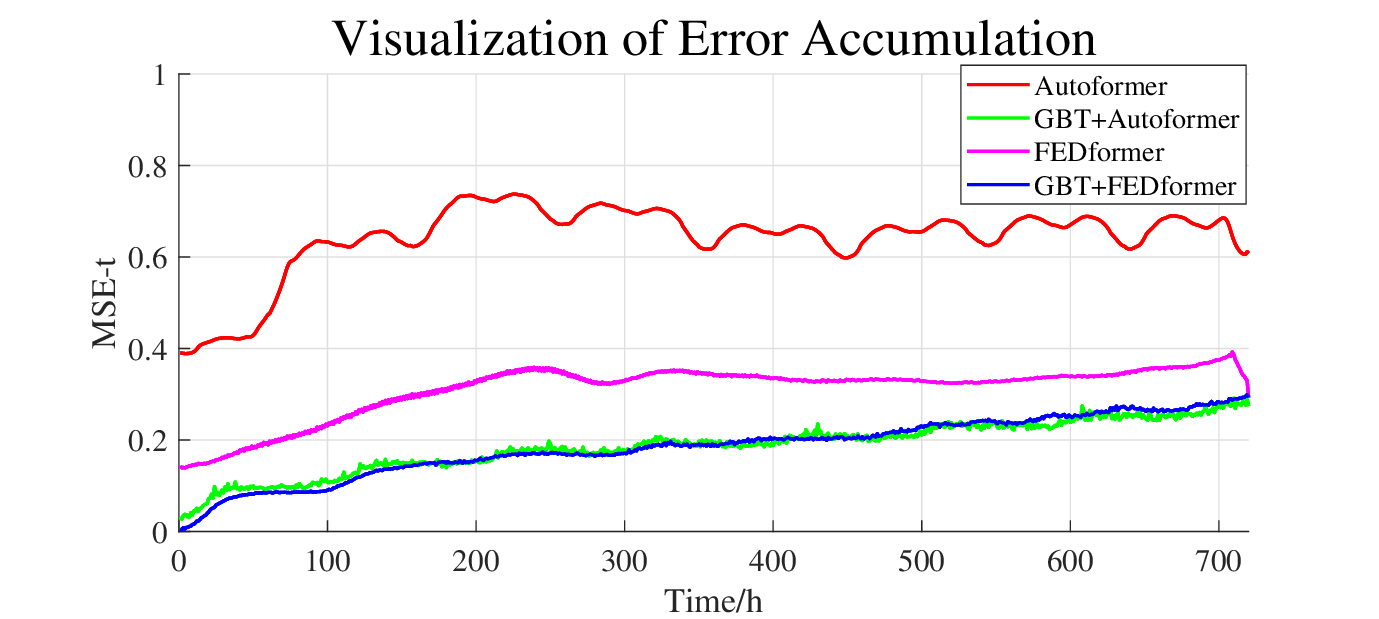}
	\caption{Forecasting error for each time step. We visualize the $MSE-t$ of baselines with/without GBT under univariate forecasting of ETTm$_2$. The prediction length is 720.}
	\label{fig5}
\end{figure}
\begin{figure*}[pos=b]
	\centering	
	\includegraphics[width=1\columnwidth]{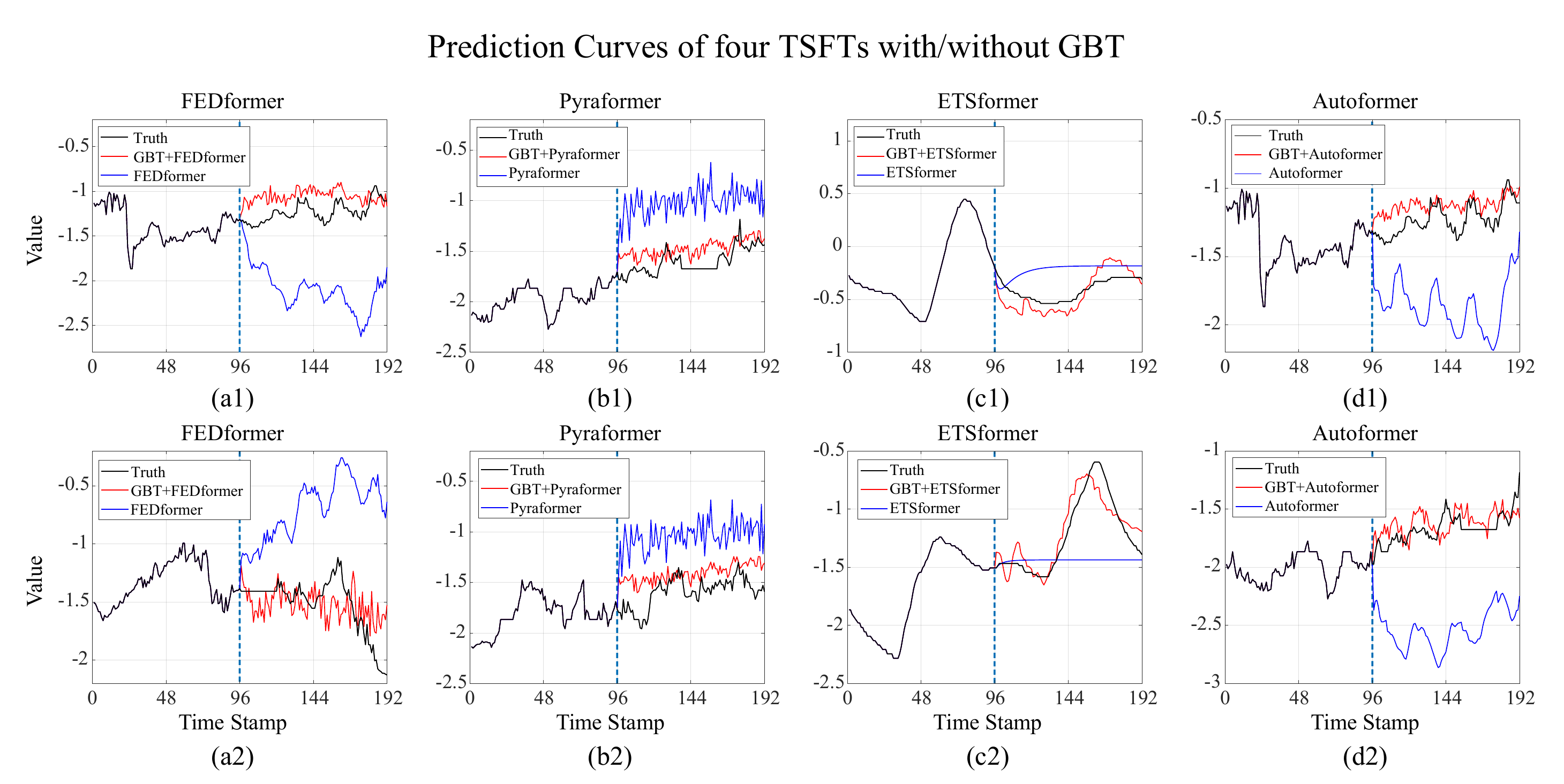}
	\caption{{Several prediction curves of four TSFTs with/without GBT. Left sequences of bule dashed lines are input sequence while right ones are prediction sequences of TSFTs with/without GBT (red/blue lines) and corresponding truths (black lines).}}	
	\label{figa4}
\end{figure*}

We visualize the forecasting error for each time step in Figure \ref{fig5} to compare extents of error accumulation among TSFTs with/without GBT. FEDformer and Autoformer are chosen as baselines and are experimented under the univariate forecasting of ETTm$_2$. We take the longest prediction length within main results, i.e., 720. The error at the $ t $-th time step is computed as $MSE-t$ which is the MSE of $ t $-th time step among all prediction windows. Figure \ref{fig5} illustrates that the $ MSE-t $ of GBT+FEDformer/GBT+Autoformer is much lower than $ MSE-t $ of FEDformer/Autoformer. Specially, $MSE-1$ of Autoformer is even bigger than the $MSE-720$ of GBT+Autoformer. This demonstrates that GBT can credibly give TSFT a `\textit{Good Beginning}'. Moreover, error accumulations of baselines are slower and more steady after combined with GBT while error accumulations of initial Autoformer and FEDformer are periodic or turbulent at the beginning/ending. These phenomena demonstrate that GBT could greatly mitigate error accumulation of TSFT.\par

{
\subsection{Showcases of GBT with TSFTs} 
To more vividly illustrate how GBT helps alleviate the over-fitting problem of TSFTs, the specific prediction curves, originating from ETTm$ _{2} $, of  several TSFTs with/without GBT are sketched in Figure \ref{figa4}. These forecasting conditions all own distribution shifts appearing between the input sequences and the prediction sequences or within the input sequences. It can be observed that these time series forecasting Transformers are prone to the over-fitting problem, barring combined with GBT. When dramatical distribution shifts occur, neither of them can even provide the prediction results whose statistics fit those of the ground truths. However, with the help of GBT, their performances become much better.\par}

{\subsection{Full Results of FEDformer and DLinear}
\label{full_result}
The full results of \{DLinear-I, DLinear-S\} and \{FEDformer-f, FEDformer-w\} are shown in Table \ref{tab9} and \ref{tab10}. Results of GBT-Vanilla is also presented for a vivid comparison.  Obviously, the best results of their two versions are better than the {single ones}, however, GBT-Vanilla still outperforms them in the major conditions. GBT-Vanilla yields relatively 22.6\%/21.5\%/15.5\%/15.1\% MSE reduction during multivariate forecasting and 33.8\%/33.8\%/ 23.4\%/21.4\% MSE reduction during univariate forecasting, when compared with DLinear-I/DLinear-S/FEDformer-f/ FEDformer-w.\par}

\setlength{\parskip}{2em} 
\begin{table}[pos=h]
	\footnotesize   
	\renewcommand{\arraystretch}{1.1}
	\centering
	\setlength\tabcolsep{4pt}
	\renewcommand{\multirowsetup}{\centering}
	\caption{Full results of FEDformer and DLinear during multivariate forecasting}
	\label{tab9}
	\begin{tabular}{c|c|cccc|cccc|cccc}
		\toprule[1.5pt]
		\multirow{2}{*}{Methods}     & \multirow{2}{*}{Metrics} & \multicolumn{4}{c|}{ETTh$ _{1} $}       & \multicolumn{4}{c|}{ETTm$ _{2} $}       & \multicolumn{4}{c}{ECL}\\
		\cmidrule(lr){3-6}
		\cmidrule(lr){7-10}
		\cmidrule(lr){11-14}
		&   & 96& 192& 336& 720& 96& 192& 336& 720& 96& 192& 336& 720\\
		\midrule[1pt]
		\multirow{2}{*}{GBT-Vanilla} & MSE & \textit{\textbf{0.398}} & \textit{\textbf{0.448}} & \textit{\textbf{0.497}} & \textit{\textbf{0.538}} & \textit{\textbf{0.189}} & \textit{\textbf{0.249}} & \textit{\textbf{0.324}}& \textit{\textbf{0.395}} & \textit{\textbf{0.143}} & \textit{\textbf{0.175}} & \textit{\textbf{0.197}} & \textit{\underline{0.235}}\\
		& MAE & \textit{\textbf{0.418}} & \textit{\textbf{0.442}} & \textit{\textbf{0.470}} & \textit{\textbf{0.505}} & \textit{\textbf{0.276}} & \textit{\textbf{0.324}}& \textit{\underline{0.368}}& \textit{\underline{0.419}}& \textit{\textbf{0.246}} & \textit{\textbf{0.277}} & \textit{\textbf{0.298}} & \textit{\textbf{0.336}} \\
		\multirow{2}{*}{DLinear-I}   & MSE                     & 0.441 & 0.492 & 0.534 & \textit{\underline{0.553}} & 0.253 & 0.435 & 0.608 & 0.977 & 0.220 & 0.219 & 0.232 & 0.268 \\
		& MAE                     & \textit{\underline{0.443}} & \textit{\underline{0.472}}& \textit{\underline{0.493}} & \textit{\underline{0.527}} & 0.334 & 0.436 & 0.516 & 0.658 & 0.317 & 0.320 & 0.333 & 0.362 \\
		\multirow{2}{*}{DLinear-S} & MSE & 0.431 & 0.474 & \textit{\underline{0.518}} & 0.560 & \textit{\underline{0.199}} & 0.287 & 0.387 & 0.544 & 0.246 & 0.246 & 0.260 & 0.294 \\
		& MAE & 0.451 & 0.479 & 0.508 & 0.559 & 0.395 & 0.363 & 0.429 & 0.506 & 0.345 & 0.348 & 0.361 & 0.388\\
		\multirow{2}{*}{FEDformer-f} & MSE                     & \textit{\underline{0.415}} & 0.474 & 0.535 & 0.680 & 0.203 & \textit{\underline{0.269}} & \textit{\underline{0.325}} & \textit{\underline{0.421}} & 0.193 & 0.201 & 0.214 & 0.246 \\
		& MAE                     & 0.453 & 0.493 & 0.524 & 0.593 & \textit{\underline{0.287}} & \textit{\underline{0.328}} & \textit{\textbf{0.366}} & \textit{\textbf{0.415}} & 0.308 & 0.315 & 0.329 & 0.355 \\
		\multirow{2}{*}{FEDformer-w} & MSE                     & 0.423 & \textit{\underline{0.448}} & 0.525 & 0.691 & 0.204 & 0.316 & 0.359 & 0.433 & \textit{\underline{0.183}} & \textit{\underline{0.195}} & \textit{\underline{0.212}} & \textit{\textbf{0.231}} \\
		& MAE                     & 0.464 & 0.473 & 0.522 & 0.618 & 0.288 & 0.363 & 0.387 & 0.432 & \textit{\underline{0.297}} & \textit{\underline{0.308}} & \textit{\underline{0.313}} & \textit{\underline{0.343}} \\
		\midrule
		\midrule
		\multirow{2}{*}{Methods}     & \multirow{2}{*}{Metrics} & \multicolumn{4}{c|}{WTH}& \multicolumn{4}{c|}{Traffic}     & \multicolumn{4}{c}{Exchange}    \\
		\cmidrule(lr){3-6}
		\cmidrule(lr){7-10}
		\cmidrule(lr){11-14}
		&   & 96& 192& 336& 720& 96& 192& 336& 720& 96& 192& 336& 720\\
		\midrule[1pt]
		\multirow{2}{*}{GBT-Vanilla} & MSE & \textit{\textbf{0.434}} & \textit{\textbf{0.481}} & \textit{\textbf{0.514}} & \textit{\textbf{0.523}} & \textbf{\textit{0.509}}& \textbf{\textit{0.520}}& \textbf{\textit{0.535}}& \textbf{\textit{0.575}}& \textbf{\textit{0.110}}& \textit{\textbf{0.179}} & \textit{\underline{0.358}}& \textit{\textbf{0.756}} \\
		& MAE & \textit{\textbf{0.466}} & \textit{\textbf{0.506}} & \textit{\textbf{0.527}} & \textit{\textbf{0.532}} & \textbf{\textit{0.282}}& \textbf{\textit{0.293}}& \textbf{\textit{0.307}}& \textbf{\textit{0.317}}& \textbf{\textit{0.249}}& \textit{\textbf{0.312}}& \textit{\underline{0.446}}& \textit{\textbf{0.655}} \\
		\multirow{2}{*}{DLinear-I}   & MSE                     & \textit{\underline{0.490}} & \underline{\textit{0.552}} & \underline{\textit{0.583}} & 0.639 & 0.794 & 0.750 & 0.756 & 0.799 & \textit{\underline{0.115}} & \textit{\underline{0.189}} & \textbf{\textit{0.279}} & \textit{\underline{0.774}} \\
		& MAE                     & \textit{\underline{0.506}} & \textit{\underline{0.548}} & \textit{\underline{0.570}} & 0.606 & 0.491 & 0.471 & 0.474 & 0.491 & \textit{\underline{0.258}} & \textit{\underline{0.335}} & \textit{\textbf{0.412}} & \textit{\underline{0.693}}\\
		\multirow{2}{*}{DLinear-S} & MSE & 0.539 & 0.592 & 0.610 & 0.653 & 0.725 & 0.665 & 0.674 & 0.716 & 0.219 & 0.350 & 0.563 & 1.076 \\
		& MAE & 0.522 & 0.557 & 0.571 & 0.600 & 0.460 & 0.438 & 0.441 & 0.457 & 0.387 & 0.481 & 0.606 & 0.799
		   \\
		\multirow{2}{*}{FEDformer-f} & MSE                     & 0.509 & 0.581 & 0.630 & \textit{\underline{0.580}} & 0.587 & 0.604 & 0.621 & 0.626 & 0.148 & 0.271 & 0.460 & 1.195 \\
		& MAE                     & 0.513 & 0.557 & 0.636 & \textit{\underline{0.586}} & 0.366 & 0.373 & 0.383 & 0.382 & 0.278 & 0.380 & 0.500 & 0.841 \\
		\multirow{2}{*}{FEDformer-w} & MSE                     & 0.553 & 0.620 & 0.661 & 0.681 & \textit{\underline{0.562}} & \textit{\underline{0.562}} & \textit{\underline{0.570}} & \textit{\underline{0.596}} & 0.139 & 0.256 & 0.426 & 1.090 \\
		& MAE                     & 0.537 & 0.571 & 0.599 & 0.607 & \textit{\underline{0.349}} & \textit{\underline{0.346}} & \textit{\underline{0.323}} & \textit{\underline{0.368}} & 0.276 & 0.369 & 0.464 & 0.800 \\
		\bottomrule[1.5pt]   
	\end{tabular}
\end{table}
\makeatletter
\setlength{\@fptop}{5pt}
\makeatother
\begin{table}
	\footnotesize   
	\renewcommand{\arraystretch}{1.1}
	\centering
	\setlength\tabcolsep{4pt}
	\renewcommand{\multirowsetup}{\centering}
	\caption{Full results of FEDformer and DLinear during univariate forecasting}
	\label{tab10}
	\begin{tabular}{c|c|cccc|cccc|cccc}
		\toprule[1.5pt]
		\multirow{2}{*}{Methods}     & \multirow{2}{*}{Metrics} & \multicolumn{4}{c|}{ETTh$ _{1} $}       & \multicolumn{4}{c|}{ETTm$ _{2} $}       & \multicolumn{4}{c}{ECL}\\
		\cmidrule(lr){3-6}
		\cmidrule(lr){7-10}
		\cmidrule(lr){11-14}
		&   & 96& 192& 336& 720& 96& 192& 336& 720& 96& 192& 336& 720\\
		\midrule[1pt]
		\multirow{2}{*}{GBT-Vanilla} & MSE& \textit{\textbf{0.051}} & \textit{\textbf{0.074}} & \textit{\textbf{0.080}} & \textit{\textbf{0.119}} & \textit{\underline{0.068}} & \textit{\textbf{0.091}} & \textit{\textbf{0.109}} & \textit{\textbf{0.163}} & \textit{\textbf{0.254}} & \textit{\textbf{0.282}} & \textit{\textbf{0.324}} & \textit{\textbf{0.359}}\\
		& MAE& \textit{\textbf{0.173}} & \textit{\textbf{0.206}} & \textit{\textbf{0.221}} & \textit{\textbf{0.276}} & \textit{\underline{0.194}} & \textit{\textbf{0.229}} & \textit{\textbf{0.257}} & \textit{\textbf{0.316}} & \textit{\textbf{0.363}} & \textit{\textbf{0.386}} & \textit{\textbf{0.417}} & \textit{\textbf{0.444}} \\
		\multirow{2}{*}{DLinear-I} & MSE & 0.111 & 0.136 & 0.166 & 0.280 & 0.094 & 0.130 & 0.164 & 0.223 & 0.411 & 0.385 & 0.410 & 0.447 \\
		& MAE & 0.258 & 0.286 & 0.325 & 0.453 & 0.237 & 0.278 & 0.316 & 0.369 & 0.473 & 0.455 & 0.470 & 0.502\\
		\multirow{2}{*}{DLinear-S} & MSE & 0.111 & 0.136 & 0.166 & 0.280 & 0.094 & 0.130 & 0.164 & 0.223 & 0.411 & 0.385 & 0.410 & 0.447 \\
		& MAE & 0.258 & 0.286 & 0.325 & 0.453 & 0.237 & 0.278 & 0.316 & 0.369 & 0.473 & 0.455 & 0.470 & 0.502 \\
		\multirow{2}{*}{FEDformer-f} & MSE & \underline{\textit{0.103}} & \underline{\textit{0.129}} & \underline{\textit{0.132}} & \textit{\underline{0.134}} & 0.072 & \textit{\underline{0.102}} & \textit{\underline{0.130}} & \textit{\underline{0.178}} & \textit{\underline{0.253}} & \textit{\underline{0.282}} & \textit{\underline{0.346}}& \textit{\underline{0.422}} \\
		& MAE & \textit{\underline{0.252}} & \textit{\underline{0.285}}& \textit{\underline{0.291}}& \textit{\underline{0.293}} & 0.206 & \textit{\underline{0.245}} & \textit{\underline{0.279}} & \textit{\underline{0.325}}& \textit{\underline{0.370}} & \textit{\underline{0.386}}& \textit{\underline{0.431}} & \textit{\underline{0.484}} \\
		\multirow{2}{*}{FEDformer-w} & MSE & 0.126 & 0.144 & 0.151 & 0.154 & \textit{\textbf{0.063}} & 0.110 & 0.147 & 0.219 & 0.262 & 0.316 & 0.361 & 0.448 \\
		& MAE & 0.279 & 0.298 & 0.299 & 0.311 & \textit{\textbf{0.189}} & 0.252 & 0.301 & 0.368 & 0.378 & 0.410 & 0.445 & 0.501\\
		\midrule
		\midrule
		\multirow{2}{*}{Methods}     & \multirow{2}{*}{Metrics} & \multicolumn{4}{c|}{WTH}& \multicolumn{4}{c|}{Traffic}     & \multicolumn{4}{c}{Exchange}    \\
		\cmidrule(lr){3-6}
		\cmidrule(lr){7-10}
		\cmidrule(lr){11-14}
		&   & 96& 192& 336& 720& 96& 192& 336& 720& 96& 192& 336& 720\\
		\midrule[1pt]
		\multirow{2}{*}{GBT-Vanilla} & MSE& \textit{\textbf{0.188}} & \textit{\textbf{0.221}} & \textit{\textbf{0.239}} & \textit{\textbf{0.218}} & \textit{\textbf{0.133}} & \textit{\textbf{0.140}} & \textit{\textbf{0.138}} & \textit{\textbf{0.174}} & \textit{\textbf{0.100}} & \textit{\textbf{0.186}} & \textit{\textbf{0.408}} & \textit{\textbf{0.925}} \\
		& MAE& \textit{\textbf{0.318}} & \textit{\textbf{0.348}} & \textit{\textbf{0.372}} & \textit{\textbf{0.349}} & \textit{\textbf{0.222}} & \textit{\textbf{0.228}} & \textit{\textbf{0.234}} & \textit{\textbf{0.268}} & \textit{\textbf{0.249}} & \textit{\textbf{0.343}} & \textbf{\textit{0.522}} & \textit{\textbf{0.743}} \\
		\multirow{2}{*}{DLinear-I} & MSE & \textit{\underline{0.207}} & \textit{\underline{0.257}}& \textit{\underline{0.293}} & 0.378 & 0.419 & 0.367 & 0.366 & 0.411 & 0.187 & 0.315 & 0.524 & \textit{\underline{0.940}} \\
		& MAE & \textit{\underline{0.336}} & \textit{\underline{0.376}} & \textit{\underline{0.402}} & 0.470 & 0.472 & 0.434 & 0.433 & 0.461 & 0.361 & 0.459 & 0.578 & \textit{\underline{0.772}}\\
		\multirow{2}{*}{DLinear-S} & MSE & 0.207 & 0.257 & 0.293 & 0.378 & 0.419 & 0.367 & 0.366 & 0.411 & 0.187 & 0.315 & 0.524 & 0.940 \\
		& MAE & 0.336 & 0.376 & 0.402 & 0.470 & 0.472 & 0.434 & 0.433 & 0.461 & 0.361 & 0.459 & 0.578 & 0.772 \\
		\multirow{2}{*}{FEDformer-f} & MSE & 0.233 & 0.291 & 0.318 & \textit{\underline{0.331}} & 0.207 & 0.205 & 0.219 & 0.244 & 0.154 & 0.286 & 0.511 & 1.301 \\
		& MAE & 0.353 & 0.406 & 0.422 & \textit{\underline{0.432}} & 0.312 & 0.312 & 0.323 & 0.344 & 0.304 & 0.420 & 0.555 & 0.879 \\
		\multirow{2}{*}{FEDformer-w} & MSE & 0.238 & 0.287 & 0.345 & 0.339 & \textit{\underline{0.170}} & \textit{\underline{0.173}}& \textit{\underline{0.178}} & \textit{\underline{0.187}} & \textit{\underline{0.131}} & \textit{\underline{0.277}}& \textit{\underline{0.426}} & 1.162 \\
		& MAE & 0.362 & 0.407 & 0.440 & 0.442 & \textit{\underline{0.263}} & \textit{\underline{0.265}} & \textit{\underline{0.266}} & \textit{\underline{0.286}} & \textit{\underline{0.284}}& \textit{\underline{0.420}} & \textit{\underline{0.511}} & 0.832\\
		\bottomrule[1.5pt]   
	\end{tabular}
\end{table}
\nolinenumbers
\clearpage
\bibliographystyle{elsarticle-harv}

\bibliography{reference}
\vskip100pt
\bio{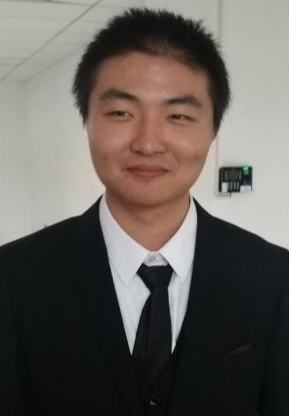}
\textbf{Li Shen} received his B.S. degree in Navigation and Control from Beihang University in China at the School of Automation Science and Electrical Engineering. He is currently completing a Doctor degree in Electronic Information Engineering at the Beihang Institute of Unmanned System. His research interests include Computer Vision and Time Series Forecasting.
\endbio
\vskip80pt
\bio{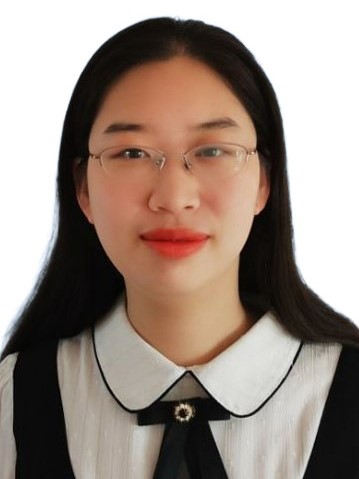}
\textbf{Yuning Wei} received her B.S. degree in Agricultural Electrification from Nanjing Agricultural University in China at the School of Engineering. She is currently completing a MA.Eng in Electronic Information Engineering at the Beihang Institute of Unmanned System. Her research interests include Deep Learning and Time Series Forecasting.
\endbio
\vskip80pt
\bio{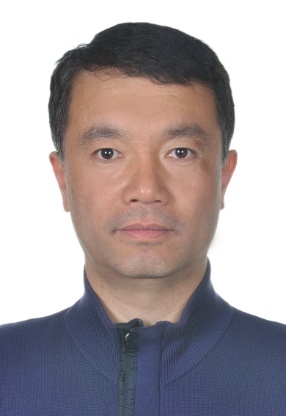}
\textbf{Yangzhu Wang} received his B.S, MA.Eng and Ph.D degrees in Measurement and Control Technology and Instrumentation Program from Beihang University in China at the Beihang Institute of Unmanned System. He is currently a researcher fellow of the Flying College of Beihang University and Beihang Institute of Unmanned System. His research interests include Computer Vision, Unmanned System and Measurement and Control Technology.
\endbio

\end{document}